\documentclass[a4paper]{article}

\usepackage{longtable}
\usepackage{tabularx}
\usepackage{graphicx}
\usepackage{amsthm}
\usepackage{amssymb}

\newcommand{\Acal}{\mathcal{A}}
\newcommand{\Bcal}{\mathcal{B}}
\newcommand{\Gcal}{\mathcal{G}}
\newcommand{\Pcal}{\mathcal{P}}
\newcommand{\Tcal}{\mathcal{T}}
\newcommand{\Vcal}{\mathcal{V}}

\newcommand{\complexes}{\mathbb{C}}
\newcommand{\prob}{\mathbb{P}}

\newtheorem{proposition}{Proposition}

\newcolumntype{C}[1]{>{\centering}p{#1}}

\begin{document}

\markboth{M. Hashimoto and R. M. Cesar Jr.}
{Object detection using keygraphs}

\title{Object Detection Using Keygraphs}

\author{Marcelo Hashimoto \and Roberto Marcondes Cesar Junior}



\maketitle

\begin{abstract}
  We propose a new framework for object detection based on a
  generalization of the keypoint correspondence framework. This
  framework is based on replacing keypoints by keygraphs,
  i.e. isomorph directed graphs whose vertices are keypoints, in order
  to explore relative and structural information. Unlike similar works
  in the literature, we deal directly with graphs in the entire
  pipeline: we search for graph correspondences instead of searching
  for individual point correspondences and then building graph
  correspondences from them afterwards. We also estimate the pose from
  graph correspondences instead of falling back to point
  correspondences through a voting table.  The contributions of this
  paper are the proposed framework and an implementation that properly
  handles its inherent issues of loss of locality and combinatorial
  explosion, showing its viability for real-time applications. In
  particular, we introduce the novel concept of keytuples to solve a
  running time issue.  The accuracy of the implementation is shown by
  results of over 800 experiments with a well-known database of
  images. The speed is illustrated by real-time tracking with two
  different cameras in ordinary hardware.

\end{abstract}

\section{Introduction}
\label{sec:introduction}

Object detection is a well-known problem in computer vision that
consists in: given a model representing an object and a scene, decide
if there are instances of the object in the scene and, if so, estimate
the pose of each instance. Solving this problem is useful for many
applications like automatic surveillance, medical analysis, and
augmented reality, but particularly challenging due to difficulties
like image distortions, background clutter, and partial occlusions.

Among the techniques for object detection available in the literature,
methods based on keypoint correspondence have been particularly
successful~\cite{lowe-1999:iccv7-2-1150,lowe-2004:ijcv-20-2-91,lf-2006:pami-28-9-1465,btv-2008:cviu-110-3-346,my-2009:siims-2-2-438}. The
outline of such methods consists in detecting interest points in the
model and the scene, selecting correspondences between the two sets of
points by analyzing the similarity between photometric descriptors,
and applying a fitting procedure over those correspondences. This
framework is not specific to object detection and has also been used
for robot navigation~\cite{moravec-1980:tech}, feature
tracking~\cite{lk-1981:ijcai7-674,tk-1991:tech,st-1994:cvpr-593},
stereo
matching~\cite{lk-1981:ijcai7-674,mcup-2002:bmvc13-384,tlf-2010:pami-32-5-815},
image retrieval~\cite{sm-1997:pami-19-5-530},
$3$D~reconstruction~\cite{btv-2008:cviu-110-3-346} and visual
odometry~\cite{akb-2008:eccv10iv-lncs-5305-102}.

In this paper, we propose a framework for object detection based on
replacing keypoints by keygraphs, i.e. isomorph directed graphs whose
vertices are keypoints, in the aforementioned outline. The motivation
for this proposal is exploring relative and structural information
that an individual point cannot provide. Because keygraphs can be more
numerous and complex, such replacement is not straightforward. Two
issues must be considered: a possible loss of locality, since large
structures might be built, and a possible combinatorial explosion,
since large sets might be enumerated. Therefore, in addition to
describing the framework itself, we describe an implementation that
takes those issues into account.

In a previous work~\cite{hc-2009:gbr7-lncs-5534-223}, the viability of
the keygraph framework has already been proved by an implementation
that uses chromaticity values for keygraph description and the Hough
transform~\cite{ballard-1981:pr-13-2-111} for pose estimation. Here,
in addition to presenting a more complete formalization and a deeper
theoretical analysis, we improve such work by using intensity values
for keygraph description and the RANSAC
algorithm~\cite{fb-1981:cacm-24-6-381} for pose estimation. The former
allowed a greater accuracy for gray-level images and the latter
allowed a wider range of possible poses.  Numerical results from
extensive experiments show that our implementation has competitive
accuracy and low running time.

This paper is organized as follows. Section~\ref{sec:related-work}
presents related works in the literature and specifies how ours
differs from them. Section~\ref{sec:overview-keygraph-framework}
presents an overview of the keygraph framework, which includes formal
definitions of keygraph and keygraph correspondence.
Section~\ref{sec:keytuples} presents the concept of keytuples, which
we introduce as a solution for a running time issue.
Section~\ref{sec:keygraph-detection} presents our implementation of
the keygraph detection, based on combinatorial and geometric criteria.
Section~\ref{sec:keygraph-description} presents our implementation of
the keygraph description, based on Fourier coefficients of intensity
profiles. Section~\ref{sec:correspondence-selection-pose-estimation}
presents our implementation of correspondence selection and pose
estimation using RANSAC.  Section~\ref{sec:experimental-results}
presents experimental results. Finally, Section~\ref{sec:conclusion}
presents our conclusions and research directions.

\section{Related work}
\label{sec:related-work}

Since the proposed keygraph framework is based on the generalization
of the original keypoint framework, as shown in
Section~\ref{sec:overview-keygraph-framework}, all work based on
keypoint correspondence may be considered related. For a literature
revision of such work, we refer to the comparative study of
descriptors by Mikolajczyk and Schmid~\cite{ms-2005:pami-27-10-1615},
the comparative study of detectors by
Mikolajczyk~et~al.~\cite{mtszmskv-2005:ijcv-65-1-43}, and the survey
by Tuytelaars and Mikolajczyk~\cite{tm-2008:ftcgv-3-3-177}.

In particular, the idea of exploring relative and structural
information in keypoint correspondence is not new. The seminal paper
of Schmid and Mohr~\cite{sm-1997:pami-19-5-530}, for example, already
recognized the importance of geometric criteria for correspondence
selection. A more recent example is the widely used shape context
descriptor~\cite{bmp-2002:pami-24-4-509}, originally defined as a
histogram of relative point positions. The difference of this paper
relies on obtaining the information from graphs with fixed size, while
the aforementioned authors consider neighborhoods with variable size.

Besides being seen as a generalization of the keypoint framework,
there are alternative interpretations for the contributions of the
keygraph framework, and the related literature varies accordingly. One
possible interpretation is that we consider a structure of parts
instead of individual parts. This idea has been successfully used in
statistical approaches for image categorization like the constellation
model proposed by Weber~et~al.~\cite{wwp-2000:eccv6i-lncs-1842-18},
the spatial priors proposed by
Crandall~et~al.~\cite{cfh-2005:cvpr-1-10}, the multiscale tree
proposed by Bouchard and Triggs~\cite{bt-2005:cvpr-1-710}, and the
sparse flexible models proposed by
Carneiro~and~Lowe~\cite{cl-2006:eccv9iii-lncs-3953-29}. The approach
of Leordeanu~et~al.~\cite{bt-2007:cvpr-1} is particularly relevant for
sharing the generic idea of using descriptors that are individually
robust and collectively discriminative. Such idea plays a role in
Section~\ref{sec:keygraph-description}.

Another, less vague, interpretation is that we use graphs whose
vertices are keypoints. We can cite the work of Tang and
Tao~\cite{tt-2005:vspets2-25}, which consists in applying graph
matching
procedures~\cite{kropatsch-2001:gbr3-210,ccls-2007:iccv11-1,kbl-2007:agtcvpr}
over graphs whose vertices are SIFT
keypoints~\cite{lowe-1999:iccv7-2-1150,lowe-2004:ijcv-20-2-91} for
detecting objects, and the work of Sirmacek and
Unsalan~\cite{su-2009:grs-47-4-1156}, which consists in applying graph
cut procedures also over graphs whose vertices are SIFT keypoints for
detecting urban areas in aerial photographs.  Those approaches,
however, are global: the objective is obtaining a single large
graph. Our approach, in contrast, is local: the objective is obtaining
multiple small graphs.

Considering such locality, the papers of Tell and
Carlsson~\cite{tc-2000:eccv6i-lncs-1842-814,tc-2002:eccv7i-lncs-2350-68}
and Kanazawa and Uemura~\cite{ku-2006:bmvc17-267} are closer. The
former not only introduced the idea of computing descriptors from
intensity profiles of keypoint pairs, which we use in
Section~\ref{sec:keygraph-description}, but also showed the importance
of relative positioning, which we use in
Section~\ref{sec:keygraph-detection}. The latter extended those
contributions using triplet vector descriptors and normalized
triangular region vectors. The pair correspondences of Tell and
Carlsson and the triplet correspondences of Kanazawa and Uemura can be
considered precursors of keygraph correspondences, but their
usefulness for pose estimation is limited by a voting table that
transforms them into an ordinary set of independent keypoint
correspondences. The Natural 3D Markers~(N3M) proposed by
Hinterstoisser~et~al.~\cite{hbn-2007:iccv11-1} do not have this
limitation and are perhaps the closest concept to keygraphs currently
available in the literature. However, Hinterstoisser~et~al. do not
analyze Natural 3D Markers directly like we do with keygraphs:
instead, they select independent keypoint correspondences and verify
which N3M correspondences are implied by those.

Structures of keypoints have also been successful in content-based
image retrieval, particularly in works based on visual words
\cite{sz-2006:lncs-4170-127,pcisz-2007:cvpr-1} where descriptors are
quantized into a visual vocabulary using clustering. This process is
also conceptually related to Section~\ref{sec:keygraph-description}:
the clustering can be interpreted as simplifying the descriptors,
while the structures can be interpreted as an attempt to use
collectivity for recovering distinctiveness lost in
quantization. Relevant examples are the doublets proposed by
Sivic~et~al.~\cite{srezf-2005:iccv10-2-370}, the visual phrases
proposed by Yuan~et~al.~\cite{ywy-2007:cvpr-1}, the TSR proposed by
Punitha and Guru~\cite{pg-2008:pr-41-6-2068}, and the $\Delta$-TSR
proposed by Hoang~et~al.~\cite{hgrm-2010:pr-43-9-3013}. Another
successful approach based on introducing a structure to compensate for
descriptor limitations is the spatial pyramid
matching~\cite{lsp-2006:cvpr-2-2169}.

\section{Overview of the keygraph framework}
\label{sec:overview-keygraph-framework}

The following table illustrates the analogy between the existing
keypoint framework and the proposed keygraph framework.

\tiny
\begin{center}
  \begin{tabular}{l|l}
    \textbf{keypoint framework} & \textbf{keygraph framework}\\
    \hline
    1 detect model keypoints~$\Pcal_m$ and scene keypoints~$\Pcal_s$ & 1 detect model keypoints~$\Pcal_m$ and scene keypoints~$\Pcal_s$\\
    & \hspace{0.15cm} detect model keygraphs~$\Gcal_m$ and scene keygraphs~$\Gcal_s$\\
    2 compute the descriptor of each keypoint & 2 compute the descriptors of each keygraph\\
    3 select keypoint correspondences & 3 select keygraph correspondences\\
    4 estimate pose using the selected correspondences & 4 estimate pose using the selected correspondences
  \end{tabular}
\end{center}
\normalsize

The numbering represents the four steps that both frameworks
analogously have: detection, description, selection, and estimation.
Keypoint detection is the first task of both frameworks, and the only
task that does not require adaptation for the keygraph framework: we
can directly use the existing algorithms.

\begin{figure}[!ht]
  \centering
  \includegraphics[scale=0.35]{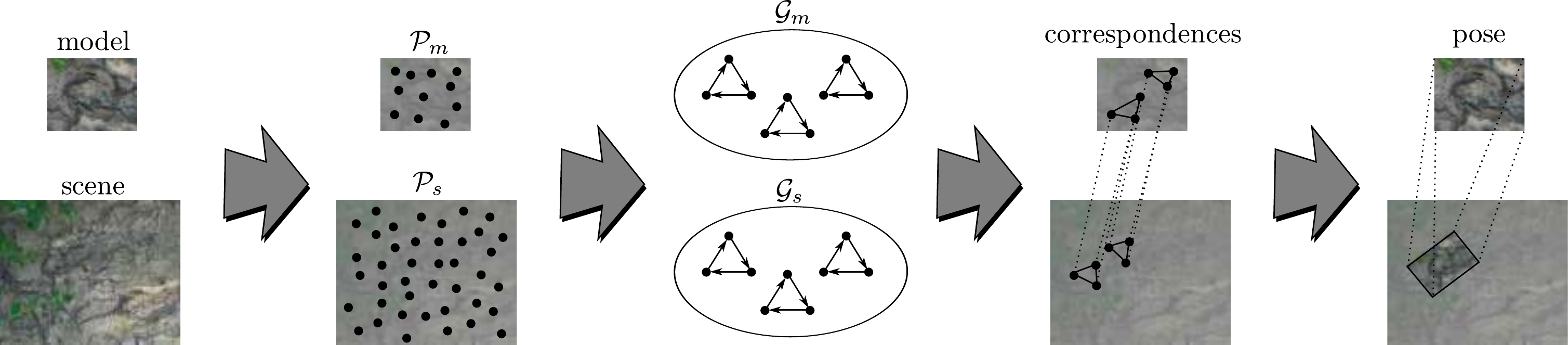}\\
  (a) \hspace{1.6cm} (b) \hspace{1.75cm} (c) \hspace{1.75cm} (d) \hspace{1.6cm} (e)
  \caption{\label{fig:overview-keygraph-framework}Illustration of the
    framework.  (a) Model and scene. (b) Detection of
    keypoints~$\Pcal_m$ and~$\Pcal_s$. (c) Detection of
    keygraphs~$\Gcal_m$ and~$\Gcal_s$. (d) Selection of
    correspondences. (e) Pose estimation.}
\end{figure}

Figure~\ref{fig:overview-keygraph-framework} illustrates the sequence
of steps defined by the proposed framework. In our implementation, a
descriptor is not global with respect to the graph: we compute local
descriptors and select correspondences by partwise analysis.

\subsection{Definition of keygraph}
\label{subsec:definition-keygraph}

As mentioned in Section~\ref{sec:introduction}, keygraphs are isomorph
directed graphs whose vertices are keypoints. More formally:
\begin{itemize}
\item each keygraph is a pair~$(\Vcal, \Acal)$, where the set of
  vertices~$\Vcal$ is a subset of keypoints and the set of
  arcs~$\Acal$ is a subset of~$\{(v, w) \colon v, w \in \Vcal
  \textrm{ and } v \not = w\}$;
\item for any keygraphs~$G_1 = (\Vcal_1, \Acal_1)$ and~$G_2 =
  (\Vcal_2, \Acal_2)$, there exists a bijection~$\iota \colon \Vcal_1
  \rightarrow \Vcal_2$ with arc preservation: $(a, b) \in \Acal_1$ if
  and only if~$(\iota(a), \iota(b)) \in \Acal_2$. Such bijection is
  called an isomorphism from~$G_1$ to~$G_2$.
\end{itemize}

The isomorphism requirement is our proposed solution for the issue of
comparability: unlike a pair of points, a pair of graphs can have
structural differences and therefore requires structural criteria for
deciding whether it is comparable. We propose bijection as a criterion
to purposefully avoid mapping multiple model keypoints to a single
scene keypoint. This contrasts, for example, with works where
super-segmentation graphs are mapped to segmentation
graphs~\cite{cbbl-2005:pr-38-11-2099}: in such works, surjections are
more adequate. The criterion of arc preservation, in its turn, avoids
mapping an arc to a pair that is not an arc. This ensures the
aforementioned partwise analysis: the isomorphism establishes which
model vertex should be compared to each scene vertex and which model
arc should be compared to each scene arc.

Note that the keygraph framework is a generalization of the keypoint
framework: a keypoint can be interpreted as a keygraph with a single
vertex, and from a single-vertex graph to another there is always an
isomorphism.

\subsection{Definition of keygraph correspondence}
\label{subsec:definition-keygraph-correspondence}

\begin{figure}[!ht]
  \centering
  \includegraphics[scale=0.35]{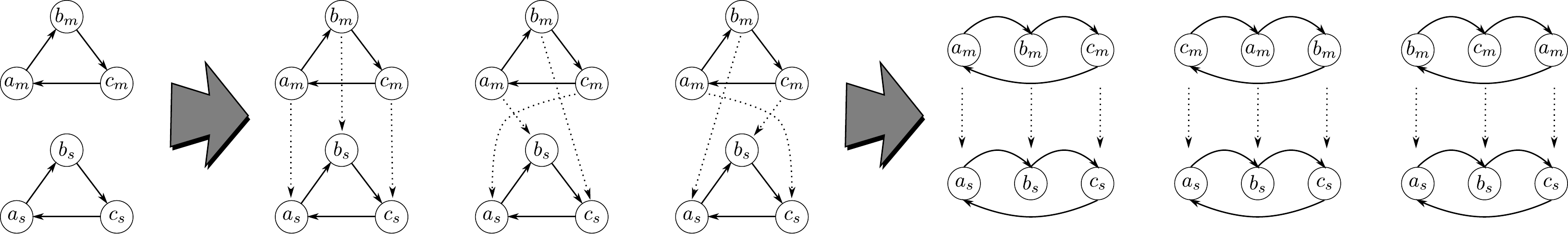}\\
  (a) \hspace{2.8cm} (b) \hspace{4.6cm} (c) \hspace{1.7cm}
  \caption{\label{fig:definition-keygraph-correspondence}Example of
    multi-comparability. (a) A model keygraph with vertices~$a_m, b_m,
    c_m$ and a scene keygraph with vertices~$a_s, b_s, c_s$. (b)
    Visual representation of the three isomorphisms from the model
    keygraph to the scene keygraph: for each isomorphism~$\iota$, a
    dashed arrow from a model vertex~$v$ to a scene vertex~$w$
    illustrates that~$\iota(v) = w$. (c) The same representation,
    after repositioning the vertices to emphasize arc preservation.}
\end{figure}

A keygraph correspondence is a triple~$(G_m, G_s, \iota)$, where~$G_m$
is a model keygraph, $G_s$ is a scene keygraph, and~$\iota$ is an
isomorphism from~$G_m$ to~$G_s$. Specifying the isomorphism is our
proposed solution for the issue of multi-comparability: from one graph
to another, more than one isomorphism can
exist. Figure~\ref{fig:definition-keygraph-correspondence} shows an
example.

In contrast, a keypoint correspondence is simply a
pair~$(p_m, p_s)$, where~$p_m$ is a model keypoint and~$p_s$ is a
scene keypoint. Note that a keygraph correspondence~$(G_m, G_s,
\iota)$ implies the set of keypoint correspondences
\begin{equation}\label{eqn:single-keygraph-correspondence}
  \{(v, \iota(v)) \colon v \textrm{ is a vertex of } G_m\}
\end{equation}
and therefore a set of keygraph correspondences~$\Gamma$ implies the
set of keypoint correspondences
\begin{equation}\label{eqn:multiple-keygraph-correspondences}
  \Pi := \bigcup_{(G_m, G_s, \iota) \in \Gamma} \{(v, \iota(v)) \colon v \textrm{ is a vertex of } G_m\}
\end{equation}
The first implication is particularly relevant for estimating pose
using RANSAC, as shown in
Section~\ref{sec:correspondence-selection-pose-estimation}, and the
second implication is particularly relevant for analyzing our
experimental results, as shown in
Section~\ref{sec:experimental-results}.

In our implementation, we assume few models and many scenes: this is
reasonable for applications such as video tracking and content-based
image retrieval. Under this assumption, detecting model keygraphs and
computing their respective descriptors once and storing them for
subsequent usage is more adequate than repeating such detection and
computation for each scene.  Therefore, we implement the following
asymmetric version of the framework.

\small
\begin{center}
  \begin{tabular}{rl}
     1 & \hspace{0.0cm} detect and store model keypoints~$\Pcal_m$\\
     2 & \hspace{0.0cm} detect and store model keygraphs~$\Gcal_m$\\
     3 & \hspace{0.0cm} for each~$G_m$ in~$\Gcal_m$\\
     4 & \hspace{0.5cm}   compute and store the descriptors of~$G_m$\\
     5 & \hspace{0.0cm} for each scene\\
     6 & \hspace{0.5cm}   detect scene keypoints~$\Pcal_s$\\
     7 & \hspace{0.5cm}   detect scene keygraphs~$\Gcal_s$\\
     8 & \hspace{0.5cm}   for each~$G_s$ in~$\Gcal_s$\\
     9 & \hspace{1.0cm}     compute the descriptors of~$G_s$\\
    10 & \hspace{1.0cm}     for each~$G_m$ in~$\Gcal_m$\\
    11 & \hspace{1.5cm}       for each isomorphism~$\iota$ from~$G_m$ to~$G_s$\\
    12 & \hspace{2.0cm}         select or reject~$(G_m, G_s, \iota)$\\
    13 & \hspace{0.0cm} estimate pose using the selected correspondences\\
  \end{tabular}
\end{center}
\normalsize

We also assume availability of time for this storage: only
correspondence selection and pose estimation must be as fast as
possible. Such assumption allows us to process the model keygraphs and
their respective descriptors to make the correspondence selection
faster. This paradigm has been successful in the literature of
keypoint correspondence, particularly in works based on indexing
structures~\cite{lowe-1999:iccv7-2-1150,lowe-2004:ijcv-20-2-91} and
machine
learning~\cite{lf-2006:pami-28-9-1465,rd-2006:eccv9i-lncs-3951-430}.

Therefore, throughout this paper we prioritize exhaustiveness
when extracting model data and prioritize speed when extracting
scene data. Such paradigm can be summarized as follows:
\begin{itemize}
\item sacrifice, for speed, the majority of scene information;
\item ensure, with exhaustiveness, that any remaining scene
  information that can be mapped to model information will be mapped
  to model information.
\end{itemize}
This is particularly relevant for keygraph detection, as seen in
Section~\ref{sec:keygraph-detection}, and also relevant in
Section~\ref{sec:keytuples}.

\section{Keytuples}
\label{sec:keytuples}

Given a set of model keygraphs and a set of scene keygraphs, obtaining
correspondences is not straightforward because a bijection from a
model graph vertex set to a scene graph vertex set might not be an
isomorphism. In Figure~\ref{fig:definition-keygraph-correspondence},
for example, the three isomorphisms represent only half of the six
possible bijections. This raises an issue:
\begin{itemize}
\item on one hand, the exhaustive verification of all bijections is
  expensive: if the keygraphs have~$k$ vertices, there are~$k!$
  possible bijections;
\item on the other hand, an isomorphism to a scene graph cannot be
  stored when only model graphs are available.
\end{itemize}

In order to solve this issue, we introduce the concept of
keytuples. Let~$G = (\Vcal, \Acal)$ be a keygraph with~$k$ vertices,
$(v_1, \ldots, v_k)$ be an ordering of~$\Vcal$, and~$\sigma$ be a $k
\times k$ binary matrix. We say~$(v_1, \ldots, v_k)$ is a keytuple
of~$G$ with respect to~$\sigma$ if
\begin{displaymath}
  \sigma_{ij} = \left\{\begin{array}{ll}
    1 & \textrm{ if } (v_i, v_j) \textrm{ belongs to } \Acal;\\
    0 & \textrm{ otherwise.}
  \end{array}\right.
\end{displaymath}
In other words, $\sigma$ establishes for each~$i, j$ whether there is
an arc in~$G$ from the $i$-th to the $j$-th vertex of the keytuple.
In Figure~\ref{fig:definition-keygraph-correspondence}, for example,
the orderings~$(a_m, b_m, c_m), (c_m, a_m, b_m), (b_m, c_m, a_m)$ are
keytuples of the model graph with respect to
\begin{displaymath}
  \left[\begin{array}{ccc}
    0 & 1 & 0\\
    0 & 0 & 1\\
    1 & 0 & 0
    \end{array}
  \right]
\end{displaymath}
because this matrix establishes that there are arcs from the first to
the second vertex, from the second to the third vertex, and from the
third to the first vertex.

From the definition of isomorphism and the definition of keytuple, it
is easy to prove the following proposition.

\begin{proposition}
  Let~$G_m = (\Vcal_m, \Acal_m), G_s = (\Vcal_s, \Acal_s)$ be
  keygraphs, $\sigma$ be a matrix, $(w_1, \ldots, w_k)$ be a keytuple
  of~$G_s$ with respect to~$\sigma$, and~$\Tcal$ be the set of
  keytuples of~$G_m$ with respect to~$\sigma$. Then a bijection~$\iota
  \colon \Vcal_m \rightarrow \Vcal_s$ is an isomorphism from~$G_m$
  to~$G_s$ if and only if there exists a keytuple~$(v_1, \ldots, v_k)$
  in~$\Tcal$ such that~$\iota(v_i) = w_i$ for all~$i$.
\end{proposition}

This allows a straightforward procedure for obtaining all isomorphisms
from a model keygraph~$G_m$ to a scene keygraph~$G_s$ given a
matrix~$\sigma$:
\begin{enumerate}
\item \label{item:keytuples-model} let~$\Tcal_m$ be the set of all
  keytuples of~$G_m$ with respect to~$\sigma$;
\item \label{item:keytuples-scene} let~$(w_1, \ldots, w_k)$ be a
  keytuple of~$G_s$ with respect to~$\sigma$;
\item for each keytuple~$(v_1, \ldots, v_1)$ in~$\Tcal_m$, let~$\iota$
  be the isomorphism from~$G_m$ to~$G_s$ such that~$\iota(v_i) = w_i$
  for all~$i$.
\end{enumerate}
Such procedure fits adequately into the paradigm of prioritizing
exhaustiveness when extracting model data and prioritizing speed when
extracting scene data: Step~\ref{item:keytuples-model} considers all
model keytuples, while Step~\ref{item:keytuples-scene} chooses a
single scene keytuple. The proposition ensures the choice can be
arbitrary.

\begin{figure}[!ht]
  \centering
  \includegraphics[scale=0.35]{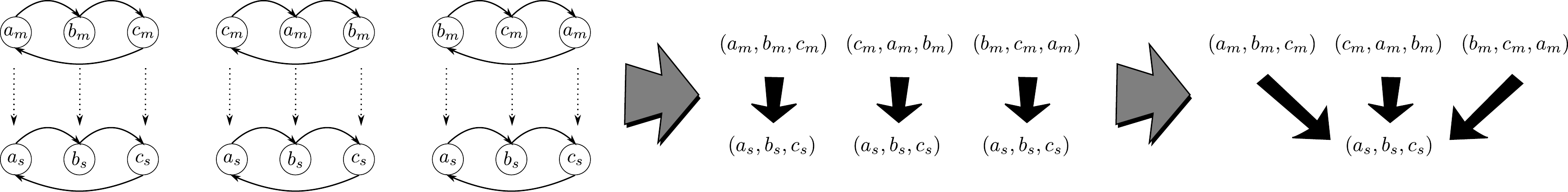}\\
  \hspace{0.8cm} (a) \hspace{4.0cm} (b) \hspace{3.1cm} (c)
  \caption{\label{fig:keytuples}Continuation of
    Figure~\ref{fig:definition-keygraph-correspondence} for an example
    of keytuples. The dashed arrows in illustration~(a) are not
    necessary because the isomorphisms are implied by vertex
    ordering. Illustration~(b) is simpler but redundant because the
    same scene keytuple is repeated. Illustration~(c) 
    emphasizes the asymmetry.}
\end{figure}

Figure~\ref{fig:keytuples} shows an example of keytuples and the
following outline adapts the asymmetric framework to them.

\small
\begin{center}
  \begin{longtable}{rl}
     1 & \hspace{0.0cm} detect and store model keypoints~$\Pcal_m$\\
     2 & \hspace{0.0cm} detect and store model keygraphs~$\Gcal_m$\\
     3 & \hspace{0.0cm} for each~$G_m$ in~$\Gcal_m$\\
     4 & \hspace{0.5cm}   compute and store the descriptors of~$G_m$\\
     5 & \hspace{0.5cm}   for each keytuple~$(v_1, \ldots, v_k)$ of~$G_m$\\
     6 & \hspace{1.0cm}     store~$(v_1, \ldots, v_k)$ in~$\Tcal_m$\\
     7 & \hspace{0.0cm} for each scene\\
     8 & \hspace{0.5cm}   detect scene keypoints~$\Pcal_s$\\
     9 & \hspace{0.5cm}   detect scene keygraphs~$\Gcal_s$\\
    10 & \hspace{0.5cm}   for each~$G_s$ in~$\Gcal_s$\\
    11 & \hspace{1.0cm}     compute the descriptors of~$G_s$\\
    12 & \hspace{1.0cm}     let~$(w_1, \ldots, w_k)$ be a keytuple of~$G_s$\\
    13 & \hspace{1.0cm}     for each keytuple~$(v_1, \ldots, v_k)$ in~$\Tcal_m$\\
    14 & \hspace{1.5cm}       let~$G_m$ be the graph in~$\Gcal_m$ such that~$(v_1, \ldots, v_k)$ is a keytuple of~$G_s$\\
    15 & \hspace{1.5cm}       let~$\iota$ be the isomorphism from~$G_m$ to~$G_s$ such that~$\iota(v_i) = w_i$ for all~$i$\\
    16 & \hspace{1.5cm}       select or reject~$(G_m, G_s, \iota)$\\
    17 & \hspace{0.0cm} estimate pose using the selected correspondences\\
  \end{longtable}
\end{center}
\normalsize

Note that this adaptation replaced a double loop in keygraphs and
isomorphisms by a single loop in keytuples.

\section{Keygraph detection}
\label{sec:keygraph-detection}

In a previous work~\cite{hc-2009:gbr7-lncs-5534-223}, the
implementation used the Good Features To Track
algorithm~\cite{st-1994:cvpr-593} to detect keypoints, as it provided
a good balance between accuracy and speed for the experiments. In this
paper, our implementation uses the
Hessian-Affine~\cite{ms-2004:ijcv-60-1-63,mtszmskv-2005:ijcv-65-1-43}
and MSER~\cite{mcup-2002:bmvc13-384} algorithms. The former is based
on extrema of Hessian determinants over multiple scales and the latter
is based on extremal regions that are maximally stable. Both are
actually covariant region detectors, converted to keypoint detectors
by taking the region centroids.

Those detectors were chosen because they provide the best results in
the aforementioned comparative study by
Mikolajczyk~et~al.~\cite{mtszmskv-2005:ijcv-65-1-43}. We decided to
use both because neither of them consistently outperforms the other in
all experiments of such study. The advantages and disadvantages of each
one influenced the experimental results presented in
Section~\ref{sec:experimental-results}.

Furthermore, following the paradigm of prioritizing exhaustiveness
when extracting model data and prioritizing speed when extracting
scene data, we use a subset of the scene keypoints. This subset is
obtained by randomly choosing a maximal set of points such that no
pair is excessively close according to the Chebyshev distance.  More
formally: given two keypoints~$(x_1, y_1)$ and~$(x_2, y_2)$,
\begin{displaymath}
  \max\{|x_2 - x_1|, |y_2 - y_2|\}
\end{displaymath}
cannot be lower than a certain value. In our experiments, the adopted
value is~$10$ pixels.

The keygraph detector is based on combinatorial and geometric
criteria. Most of such criteria is applied to both model graphs and
scene graphs, but some of them are applied only to the scene.  We
start by describing the shared criteria and conclude the section with
the criteria specific to the scene.

\begin{figure}[!ht]
  \centering
  \begin{tabular}{C{4cm}C{4cm}C{4cm}}
    \includegraphics[scale=0.5]{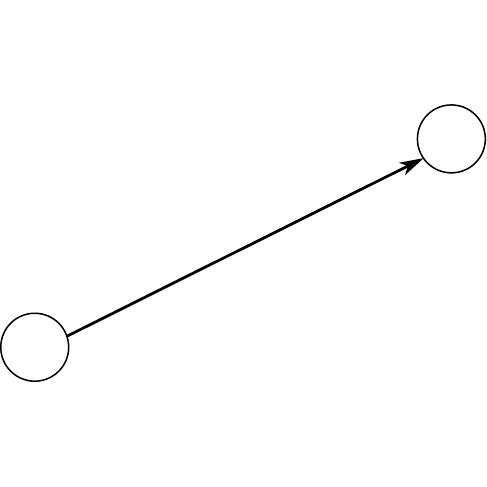}&
    \includegraphics[scale=0.5]{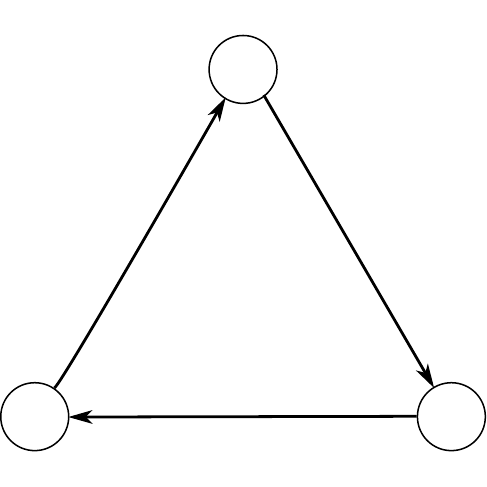}&
    \includegraphics[scale=0.5]{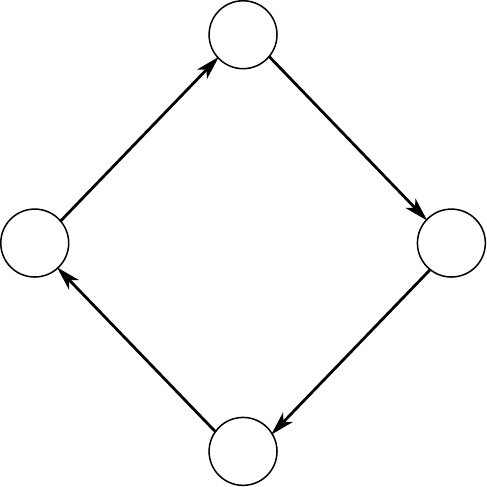}\tabularnewline[0.1cm]
    $\left[\begin{array}{cc}
      0 & 1\\
      0 & 0
      \end{array}
    \right]$&
    $\left[\begin{array}{ccc}
      0 & 1 & 0\\
      0 & 0 & 1\\
      1 & 0 & 0
      \end{array}
    \right]$&
    $\left[\begin{array}{cccc}
      0 & 1 & 0 & 0\\
      0 & 0 & 1 & 0\\
      0 & 0 & 0 & 1\\
      1 & 0 & 0 & 0
      \end{array}
    \right]$\tabularnewline[1.0cm]
    (a) & (b) & (c)
  \end{tabular}
  \caption{\label{fig:keygraph-detection-combinatorial}The three
    structures chosen and their respective matrices. 
    (a) Two vertices and the arc from
    one to the other. (b) Circuit with three vertices. (c) Circuit
    with four vertices.}
\end{figure}

The combinatorial criterion is the structure of the
keygraphs. Figure~\ref{fig:keygraph-detection-combinatorial} shows the
three structures chosen for our experiments, and the respective
matrices chosen for defining keytuples.

Such structures and matrices were chosen because obtaining keytuples
from them is straightforward and fast: for the first structure the
keytuple is unique, and for the second and third structures a keytuple
is always a cyclic permutation of another keytuple. Larger circuits
could have been used, but the experimental results of
Section~\ref{sec:experimental-results} suggested that the possible
gain in accuracy would not justify the loss in running time.

The geometric criteria, on the other hand, are three. The first
criterion is minimum distance: two vertices of a keygraph cannot be
excessively close according to the Chebyshev distance. This criterion
was chosen because, as shown in
Section~\ref{sec:keygraph-description}, we compute descriptors from
intensity profiles. An intensity profile is the sequence of values of
the pixels intersected by the straight line between the two vertices
of an arc. Our implementation uses the Bresenham
algorithm~\cite{bresenham-1965:ibmsj-4-1-25} to obtain those
values. Since the length of a profile obtained with such algorithm is
the Chebyshev distance between the respective vertices, the criterion
avoids short arcs and therefore poor descriptors. In our experiments,
we use a minimum distance of~$10$ pixels chosen empirically. The same
distance is used to sample the scene keypoints.

The second geometric criterion is maximum distance. This criterion was
chosen to avoid long arcs and therefore the consequences of loss of
locality: short intensity profiles are more robust to perspective
transforms and more adequate for non-planar objects. For the model
keygraphs, we use a maximum distance of~$100$ pixels chosen
empirically. For the scene keygraphs, we do not use a maximum distance
because in our experiments the instance of the object in the scene can
be much larger than the model.

Finally, the third criterion is relative positioning: if the graph is
one of the two circuits, its arcs must have clockwise direction. This
criterion was chosen because we assume that mirroring is not one of
the possible distortions that an instance of the object can suffer in
the scene. Under this assumption, mapping a clockwise circuit to a
counter-clockwise circuit does not make sense. Deciding whether a
circuit is clockwise requires a straightforward and fast cross-product
operation.

\begin{figure}[!ht]
  \centering
  \begin{tabular}{C{4cm}C{4cm}C{4cm}}
    \includegraphics[scale=0.4]{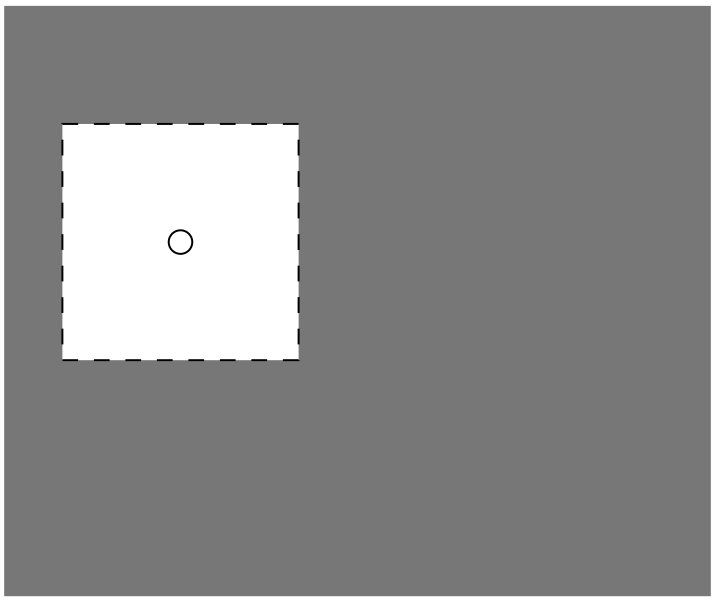}&
    \includegraphics[scale=0.4]{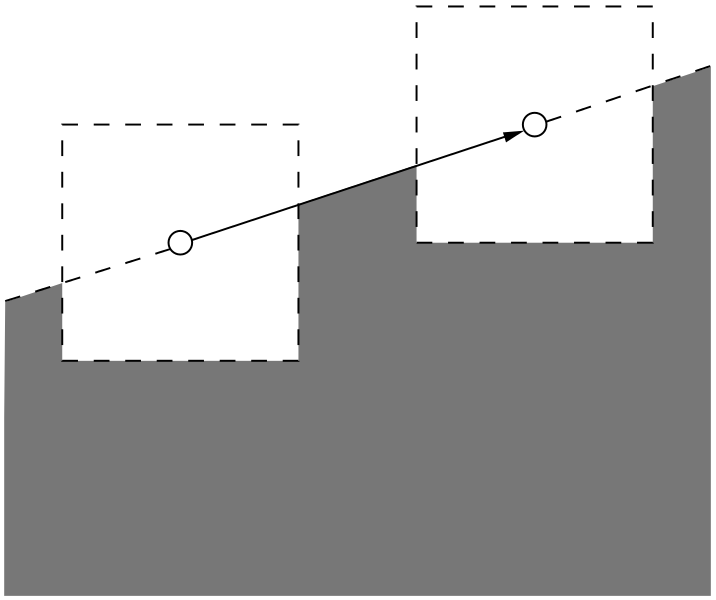}&
    \includegraphics[scale=0.4]{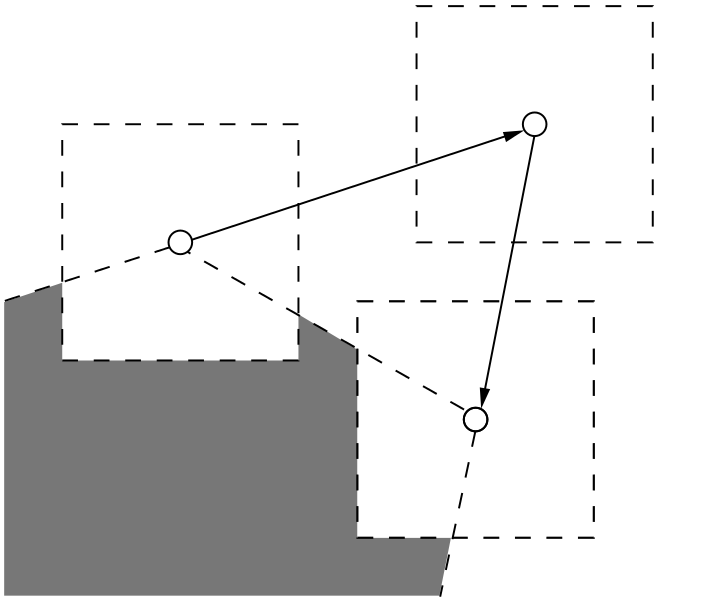}\tabularnewline
    (a) & (b) & (c)
  \end{tabular}
  \caption{\label{fig:keygraph-detection-geometric}Potential of
    geometric criteria for reducing the number of keygraphs.
    Specifically, minimum distance and relative positioning.  (a)
    After choosing the first vertex, the gray set of points that can
    be the second vertex is limited by minimum distance. (b) After
    choosing the second one, the set of points that can be the third
    one is limited by minimum distance and relative positioning. Note
    that more than half of the Euclidean plane does not belong to the
    set at this stage. (c) After choosing the third, the set that can
    be the fourth is limited even further.}
\end{figure}

Note that the geometric criteria can alleviate the combinatorial
explosion by reducing the number of graphs, as illustrated by
Figure~\ref{fig:keygraph-detection-geometric}.

\begin{figure}[!ht]
  \centering
  \begin{tabular}{cccc}
  \includegraphics[width=1.5cm]{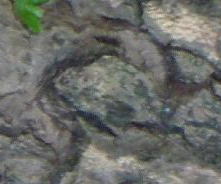}&
  \includegraphics[width=3cm]{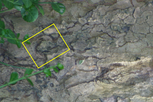}&
  \includegraphics[width=3cm]{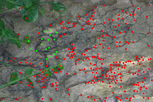}&
  \includegraphics[width=3cm]{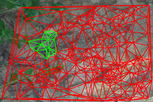}\\
  (a) & (b) & (c) & (d)
  \end{tabular}
  \caption{\label{fig:keygraph-detection-delaunay}Example of Delaunay
    triangulation (a) Model. (b) Scene and pose of the instance of the
    object in the scene. (c) Scene keypoints: points that can be
    mapped to the model are emphasized. (d) Delaunay triangulation of
    the keypoints: edges and faces that can be mapped to the model are
    emphasized.}
\end{figure}

\begin{figure}[!ht]
  \centering
  \begin{tabular}{C{4cm}C{4cm}C{4cm}}
  \includegraphics[width=2.5cm]{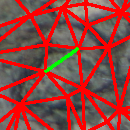}&
  \includegraphics[width=2.5cm]{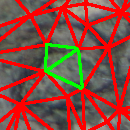}&
  \includegraphics[width=2.5cm]{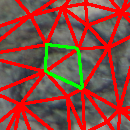}\tabularnewline
  (a) & (b) & (c)
  \end{tabular}
  \caption{\label{fig:keygraph-detection-internal}Obtaining a circuit
    with four vertices from an internal edge of a Delaunay
    triangulation.  (a) The internal edge. (b) The two faces that
    share the internal edge.  (c) The symmetric difference of the edge
    sets of the two faces.}
\end{figure}

From a theoretical point of view, the aforementioned combinatorial and
geometric criteria are enough to keep the number of model keygraphs
tractable. Since keygraphs are isomorph, the chosen structures ensure
that such number is either~$O(n^2)$, $O(n^3)$, or~$O(n^4)$, where~$n$
is the number of keypoints. In our implementation, we store the
descriptors in an indexing structure that allows correspondence
selection in expected logarithmic time. Therefore, since~$\log kn = k
\log n$, correspondence selection is sublinear on the keypoints. A
superlinear number of scene keygraphs, however, is not
tractable. Therefore, for those keygraphs we choose an additional
geometric criterion: they must be obtained from the edges and faces of
the Delaunay triangulation~\cite{dcvo-2008:cgaa3} of the keypoints.
Figure~\ref{fig:keygraph-detection-delaunay} shows an
example. Obtaining the first structure from edges and the second
structure from faces is straightforward. The third structure can be
obtained from internal edges, as illustrated by
Figure~\ref{fig:keygraph-detection-internal}.

Since the Delaunay triangulation is a planar graph, the
Euler formula can be used to prove that the number of edges and
faces is $O(n)$~\cite{dcvo-2008:cgaa3}.  Furthermore, such
triangulation can be obtained in~$O(n \log n)$
time~\cite{fortune-1986:ascg2-313}.

Note that for dense keypoint detectors, such as Hessian-Affine, the
Delaunay triangulation can have an excessive number of short edges and
therefore an excessive reduction on the number of keygraphs. This
explains why we sample the scene keypoints according to the minimum
distance. Note also that, although the triangulation is not robust to
perspective transforms, it does not have to be: we do not care if the
entire planar graph can me mapped to the model keygraphs, only if each
scene keygraph individually can. Again, we prioritize exhaustiveness
when extracting model data and prioritize speed when extracting scene
data.

\section{Keygraph description}
\label{sec:keygraph-description}

In our implementation, as shown in
Section~\ref{sec:overview-keygraph-framework}, we select
correspondences by partwise analysis and the isomorphism establishes
which model vertex should be compared to each scene vertex and which
model arc should be compared to each scene arc. The descriptor chosen
for the vertices is the widely used SIFT
descriptor~\cite{lowe-1999:iccv7-2-1150,lowe-2004:ijcv-20-2-91}, which
is based on a weighted histogram of gradient orientations and provides
the best results in the comparative study of Mikolajczyk and
Schmid~\cite{ms-2005:pami-27-10-1615}. The descriptor chosen for the
arcs is based on Fourier coefficients of the intensity profile and is
inspired by Tell and
Carlsson~\cite{tc-2000:eccv6i-lncs-1842-814,tc-2002:eccv7i-lncs-2350-68}:
given an intensity profile~$f \colon \{0, \ldots, l - 1\} \rightarrow
\{0, \ldots, 255\}$, its discrete Fourier transform is the function~$F
\colon \{0, \ldots, l - 1\} \rightarrow \complexes$ defined by
\begin{displaymath}
  F(x) = \sum^{l - 1}_{y = 0} e^{-\frac{2 \pi i}{l} xy} f(y)
\end{displaymath}
and its descriptor is defined by
\begin{displaymath}
  (a_1, b_1, \ldots, a_m, b_m) \textrm{ where } F(x) = a_x + b_x i.
\end{displaymath}
Choosing only the lowest-frequency coefficientes ensures both a
shorter descriptor size and a greater robustness to noise.

Intensity profiles are naturally invariant to rotation. They are not
invariant to perspective transforms, but are sufficiently robust under
the maximum distance criterium shown in
Section~\ref{sec:keygraph-detection}. Discarding~$F(0)$ ensures
invariance to brightness and dividing by
\begin{displaymath}
  \sqrt{\sum^{m}_{x = 1} (a^2_x + b^2_x)}
\end{displaymath}
ensures invariance to contrast. In our experiments, we have chosen~$m
= 3$ empirically and therefore an arc descriptor has~$6$ dimensions.

Profile-based descriptors are inherently poorer than region-based
descriptors, but also have inherent advantages.
\begin{itemize}
\item \textbf{Fast computation.} SIFT is widely recognized as one of
  the most accurate descriptors in the literature, but is also known
  for its slow computation. The focus of many recent
  works~\cite{btv-2008:cviu-110-3-346,akb-2008:eccv10iv-lncs-5305-102,em-2009:cvpr-9,tlf-2010:pami-32-5-815}
  is keeping the accuracy while reducing running time. This makes the
  simplicity of intensity profiles particularly interesting.
\item \textbf{Low dimensionality.} In addition to slow computation,
  SIFT also has the very high dimensionality of~$128$.  This raises
  the difficulty of building indexing structures due to the curse of
  dimensionality, the exponential increase in volume that reduces the
  significance of dissimilarity functions. In fact, SIFT descriptors
  have been used as experimental data for indexing structure
  proposals~\cite{sh-2008:cvpr-1,ml-2009:visapp4-1-331} and descriptor
  evaluation
  studies~\cite{bgrcp-2010:mcit-101,bgrcp-2010:riao9-1}. The
  aforementioned works with focus on reducing running time provide
  descriptors with lower dimensionality, and dimensionality reduction
  procedures have also been used in the
  literature~\cite{ks-2004:cvpr-2-506,ms-2005:pami-27-10-1615}. This
  makes the simplicity of intensity profile further interesting.
\item \textbf{Natural robustness.} Since regions are not naturally
  invariant to rotation like intensity profiles, invariance must be
  provided by the descriptor. SIFT and
  SURF~\cite{btv-2008:cviu-110-3-346}, for example, assign an
  orientation to each keypoint and adapt accordingly. Furthermore,
  intensity profiles provide a well-defined concept of size for
  normalization: the distance between the two vertices of the
  respective arc. In contrast, not all keypoint detectors provide such
  concept: although there are detectors that provide scale
  values~\cite{lowe-1999:iccv7-2-1150,lowe-2004:ijcv-20-2-91,btv-2008:cviu-110-3-346},
  elliptical
  regions~\cite{ms-2004:ijcv-60-1-63,mtszmskv-2005:ijcv-65-1-43}, and
  complete contours~\cite{mcup-2002:bmvc13-384}, there are also
  detectors that do not provide anything~\cite{hs-1988:avc4-147}.
\item \textbf{Weaker discrimination.} Individually, regions are more
  discriminative than profiles. However, this apparent advantage
  becomes a shortcoming in partwise analysis: if descriptors have been
  designed to be as discriminative as possible individually, they can
  be excessively discriminative collectively. For example: using the
  distance between SIFT descriptors to decide whether two single
  keypoints are similar has a high chance of success, but using the
  same criterion to compare two sets of keypoints might be
  unreasonably harsh.
\end{itemize}

\begin{figure}[!ht]
  \centering
  \begin{tabular}{C{4cm}C{4cm}C{4cm}}
    \includegraphics[width=2.5cm]{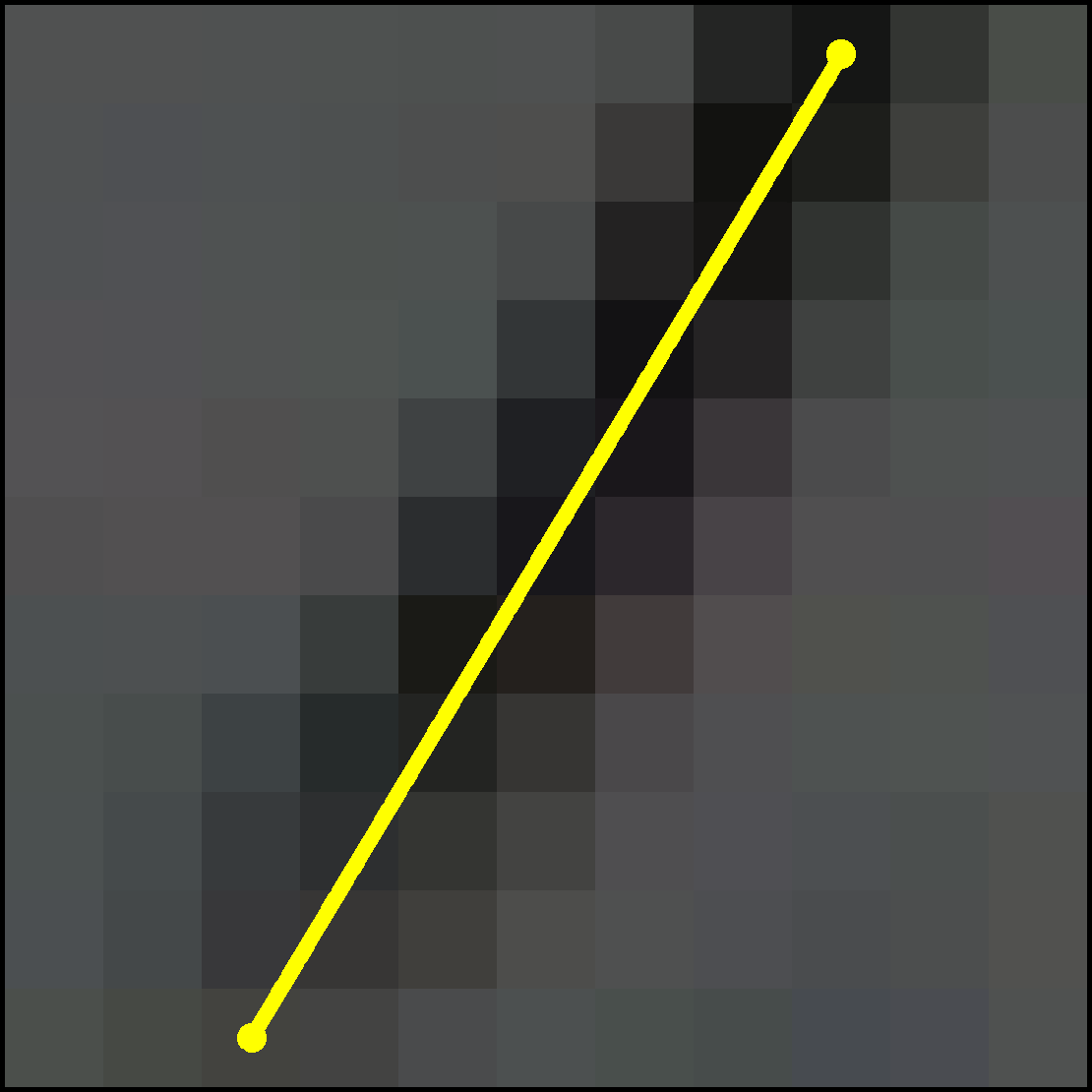}&
    \includegraphics[width=2.5cm]{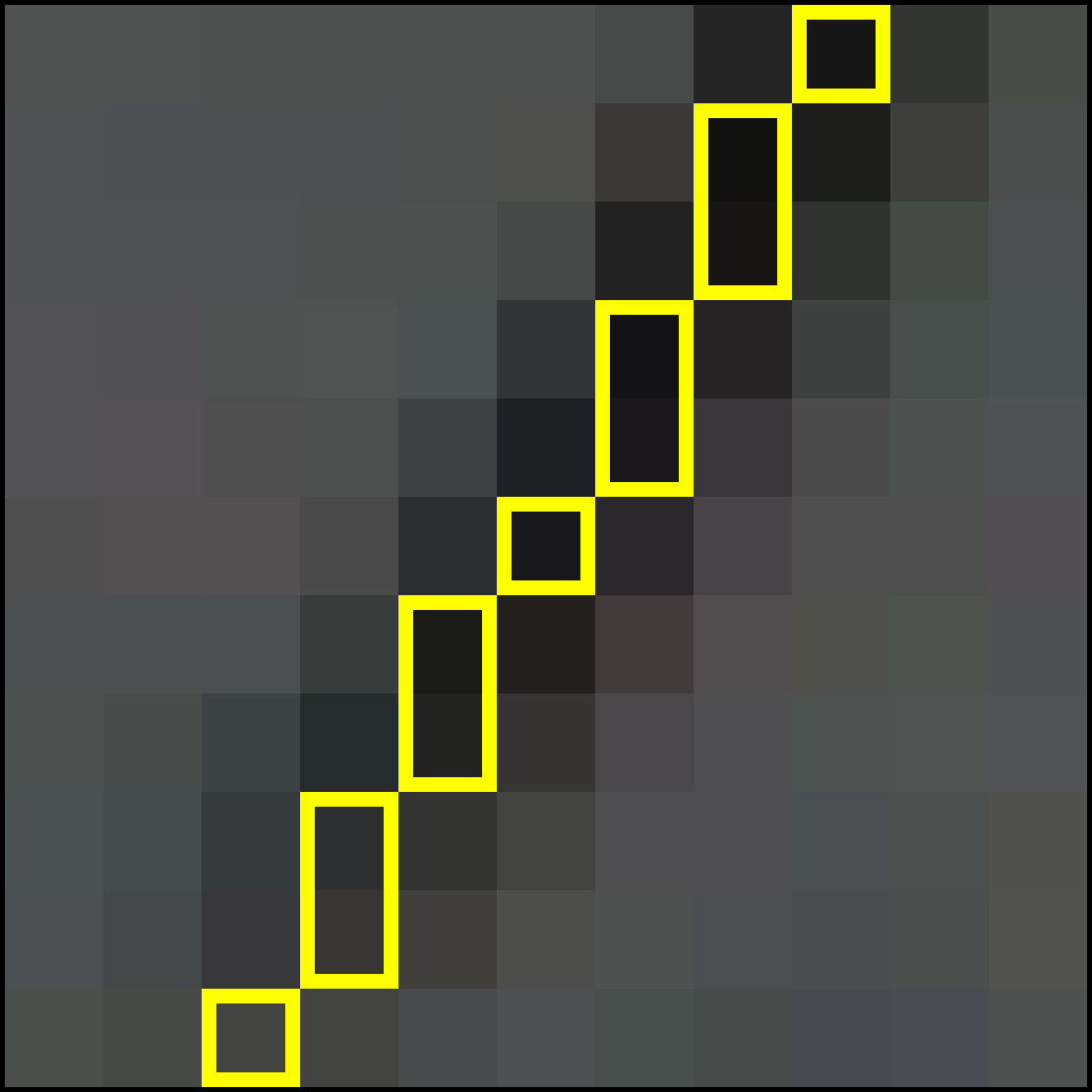}&
    \includegraphics[width=2.5cm]{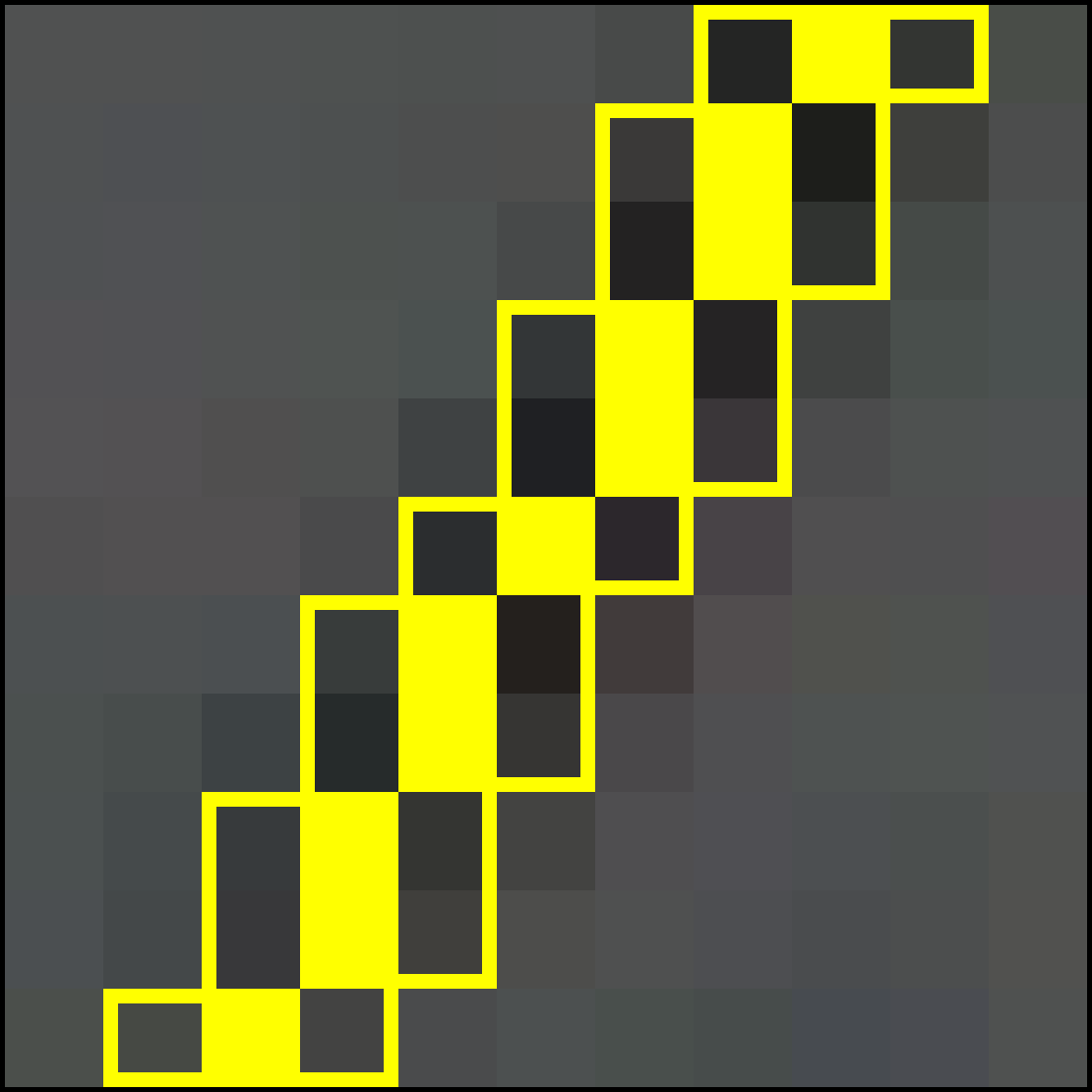}\tabularnewline
    (a) & (b) & (c)
  \end{tabular}
  \caption{\label{fig:keygraph-description}Mean of intensity
    profiles. (a) The straight line between the two vertices of an
    arc. (b) The sequence of pixels intersected by the line. (c) The
    same pixels and one additional parallel sequence on each side: the
    Fourier transform is computed from a scaled mean of the three
    sequences.}
\end{figure}

For additional robustness, we apply three smoothing procedures over
the intensity profiles before computing the Fourier transform. First,
a Gaussian blur is applied to the image.  Then, the mean of parallel
intensity profiles is computed, as illustrated by
Figure~\ref{fig:keygraph-description}.  Finally, this mean is scaled
to a length~$l$ with linear interpolation.  In our experiments, we
have empirically chosen~$l = 30$.

\section{Correspondence selection and pose estimation}
\label{sec:correspondence-selection-pose-estimation}

\begin{figure}[!ht]
  \centering
  \includegraphics[scale=0.65]{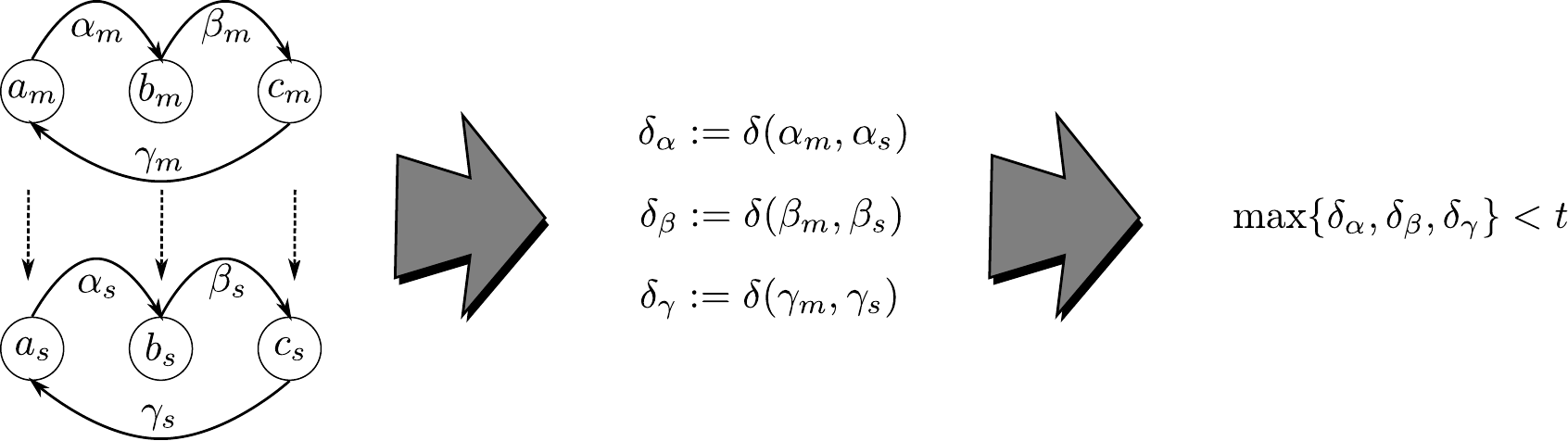}\\[0.1cm]
  (a) descriptors \hspace{1.8cm} (b) dissimilarity \hspace{1.9cm} (c) threshold \hspace{0.2cm}
  \caption{\label{fig:correspondence-selection}Example of keygraph
    correspondence selection using arc dissimilarity thresholding. (a)
    An isomorphism from a model keygraph with arc
    descriptors~$\alpha_m, \beta_m, \gamma_m$ to a scene keygraph with
    arc descriptors~$\alpha_s, \beta_s, \gamma_s$. (b) The three arc
    dissimilarities implied by the isomorphism. (c) The thresholding
    condition for selecting a correspondence.}
\end{figure}

The procedure for selecting or rejecting a keygraph correspondence in
our implementation is a simple thresholding. For each vertex
correspondence implied by the isomorphism, a dissimilarity function is
applied to the descriptors, and the correspondence is rejected if one
of the dissimilarities is higher than a certain threshold. The same
verification is made for arcs, as illustrated by
Figure~\ref{fig:correspondence-selection}.

Note that such procedure allows more than one correspondence per model
graph or scene graph. The RANSAC-based pose estimation step is
responsible for recognizing the correspondences that have been
incorrectly selected and ignoring them. RANSAC is a probabilistic
fitting procedure, specifically designed to handle high proportions of
outliers. In general terms, it is based on iteratively choosing
random sets of correspondences: for each set, a pose candidate is
estimated and the number of correspondences that are inliers with
respect to such candidate is stored as its score. The number of
iterations is probabilistically adapted according to the best score: a
high number of inliers implies a lower expected number of iterations
before finding an optimal candidate.  The actual number eventually
reaches either this expected number or an upper bound, specified to
avoid excessive running time. If the best score is sufficiently good
when this happens, the respective candidate is chosen as the pose.

Given a set of keygraph correspondences~$\Gamma$, a straightforward
procedure for estimating the pose is applying RANSAC to the
correspondences defined in
equation~(\ref{eqn:multiple-keygraph-correspondences}) or over a set
of frequent keypoint correspondences obtained with a voting
table~\cite{tc-2000:eccv6i-lncs-1842-814,tc-2002:eccv7i-lncs-2350-68,ku-2006:bmvc17-267}. However,
both procedures discard a valuable information: if a keygraph
correspondence is correct, then all keypoint correspondences it
implies are correct simultaneously.  In our implementation, we apply
RANSAC to a family of sets as
equation~(\ref{eqn:single-keygraph-correspondence}). This reduces
significatively the expected number of RANSAC iterations, as shown in
Section~\ref{sec:experimental-results}.

\section{Experimental results}
\label{sec:experimental-results}

Our implementation was written in~C++ and based on the OpenCV
library~\cite{bk-2008:opencv}. The Delaunay triangulation was computed
with the Triangle library~\cite{shewchuk-1996:acgtge-lncs-1148-203}
and the Fourier transform was computed with the FFTW
library~\cite{fj-2005:ieee-93-2-216}. In RANSAC, the best candidate is
refined with the Levenberg-Marquardt
algorithm~\cite{levenberg-1944:qam-2-2-164,marquardt-1963:sjam-11-2-431}.

\begin{figure}[!ht]
  \centering
  \begin{tabular}{ccc}
    \includegraphics[width=1.25cm]{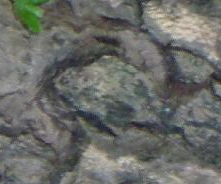}
    \includegraphics[width=2.5cm]{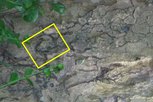}&
    \includegraphics[width=1.25cm]{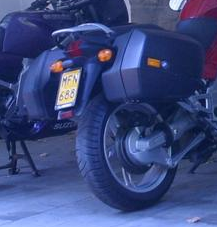}
    \includegraphics[width=2.5cm]{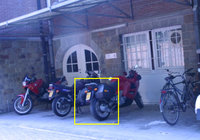}&
    \includegraphics[width=1.25cm]{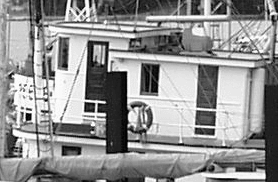}
    \includegraphics[width=2.5cm]{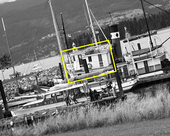}\\
    (a) bark & (b) bikes & (c) boat\\[0.1cm]
    \includegraphics[width=1.25cm]{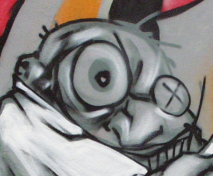}
    \includegraphics[width=2.5cm]{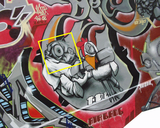}&
    \includegraphics[width=1.25cm]{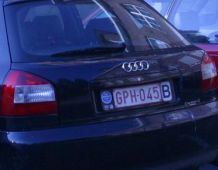}
    \includegraphics[width=2.5cm]{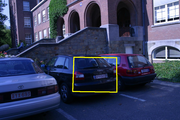}&
    \includegraphics[width=1.25cm]{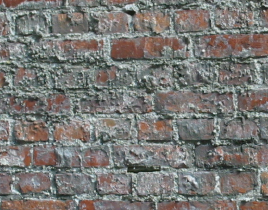}
    \includegraphics[width=2.5cm]{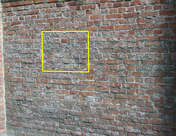}\\
    (d) graf & (e) leuven & (f) wall
  \end{tabular}
  \caption{\label{fig:experimental-results-manual-dataset}Image pairs
    used in the first experiment, with visual representations of the
    homographies. (a) bark: textured image with change of scale and
    rotation. (b) bikes: structured image with change of blur. (c)
    boat: structured image with change of scale and rotation. (d)
    graf: structured image with change of viewpoint. (e) leuven:
    structured image with change of brightness and contrast.  (f)
    wall: textured image with change of viewpoint.}
\end{figure}

In our first experiment, we used image pairs manually obtained from
the dataset shared by Mikolajczyk and
Schmid~\cite{ms-2005:pami-27-10-1615}. This dataset is partitioned
in~$8$ subsets: each subset has one model, five scenes, and five
homographies modelling the respective poses. The five scenes represent
different degrees of a specific distortion. We chose~$6$ subsets and,
from each one, obtained the scene representing the lowest degree of
distortion and a manually cropped sub-image of the
model. Figure~\ref{fig:experimental-results-manual-dataset} shows the
image pairs.

The objective of the experiment was comparing the accuracy of the
three different structures shown in
Section~\ref{sec:keygraph-detection} for different images. We were
interested in discovering the best structure for low-degree
distortions before considering high-degree distortions. The~$2$
subsets not chosen were trees, textured image with change of blur, and
ubc, structured image with change of compression. Both were discarded
for providing results excessively different: much worse for the former
and much better for the latter.

We cropped the models to bring the dataset closer to the case of
object detection in cluttered backgrounds: the cropping
significatively increases the number of scene images that have no
correspondence in the model, raising the difficulty in particular for
the textured images. The accuracy measure chosen was the
recall-precision curve: recall is the fraction of correct
correspondences that is selected, and precision is the fraction of
selected correspondences that is correct. Those two values are
computed for different thresholds and a curve is plotted accordingly.

Since we were also interested in comparing the keygraph framework with
the keypoint framework, we also plotted the precision-recall curve for
individual keypoint correspondences. The keypoint descriptor chosen to
select those correspondences was SIFT. In order to avoid bias for our
method, we established that the recall is always computed from
keypoint correspondences: in the case of keygraphs, such
correspondences are obtained from
equation~(\ref{eqn:multiple-keygraph-correspondences}).

\begin{figure}[!ht]
  \centering
  \begin{tabular}{cc}
    \includegraphics[scale=0.3]{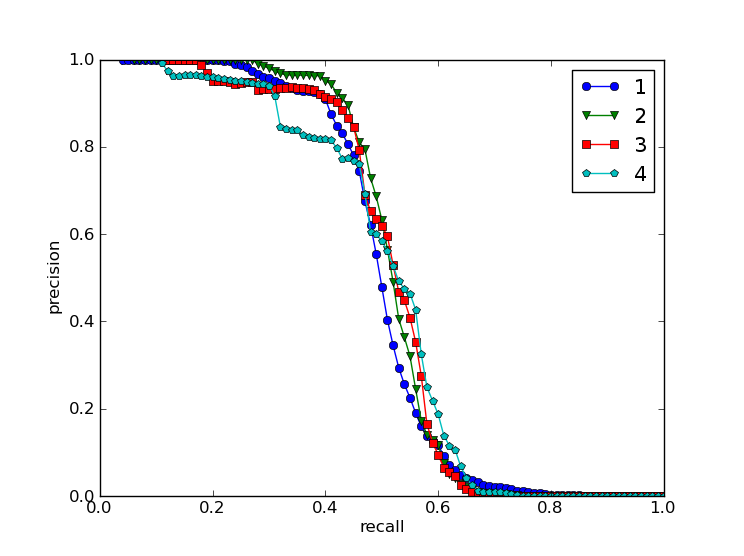}&
    \includegraphics[scale=0.3]{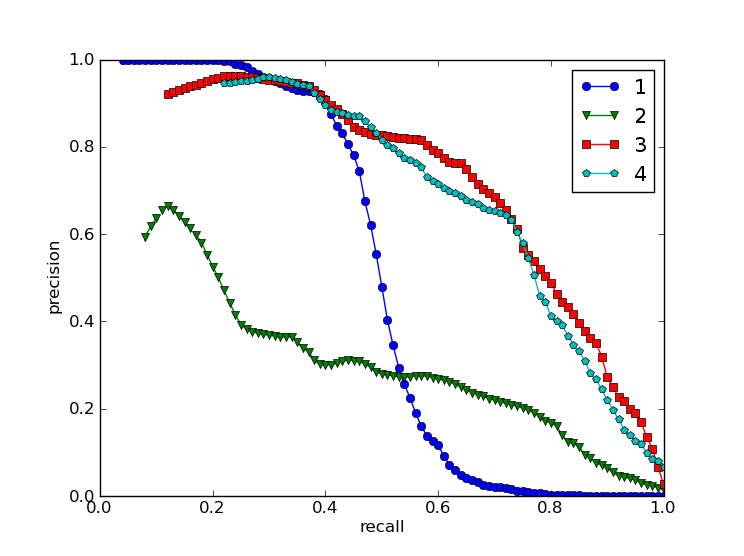}\\
    (a) vertices with SIFT & (b) arcs with Fourier
  \end{tabular}
  \caption{\label{fig:experimental-results-manual-all}Results for the
    two sequences of measurements in the first experiment: (a) for
    vertices with SIFT and (b) for arcs with Fourier. The name of each
    curve represents the number of vertices in the structure: the
    first one is the baseline curve for individual keypoint
    correspondences, hence the same in both sequences, and the other
    three are for the three structures shown in
    Section~\ref{sec:keygraph-detection}. Each curve is the mean of
    the curves across all pairs of images.}
\end{figure}

Finally, we also established that a keygraph correspondence was
correct if all its implied keypoint correspondences were correct, and
a keypoint correspondence was correct if the Euclidean distance from
the groundtruth was less than~$3$ pixels. This value was chosen
empirically. Figure~\ref{fig:experimental-results-manual-all} shows
the results for two sequences of measurements: one with SIFT
descriptors on vertices and one with Fourier descriptors on arcs.  For
both sequences the keypoint detector was the Hessian-Affine algorithm
and the dissimilarity measure was the Euclidean distance.

For vertices with SIFT, the similarity of all curves shows that
naively using a state-of-the-art keypoint descriptor in the keygraph
framework is not necessarily the best approach. For arcs with Fourier
descriptors, we observed that keygraphs consistently had worse
precisions for lower recalls, but consistently had better precisions
for higher recalls.  This was true even for the simplest structure
with two vertices, which does not have the advantages of relative
positioning, although its overall precision was much worse than the
others.  We also observed that the curves for the two circuits were
similar. This discouraged us from trying larger circuits, as the
difference in accuracy did not seem to justify the increase in running
time.

\begin{figure}
  \centering
  \includegraphics[width=1.1cm]{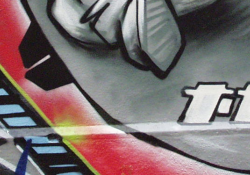}
  \includegraphics[width=1.1cm]{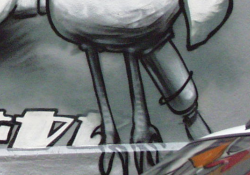}
  \includegraphics[width=1.1cm]{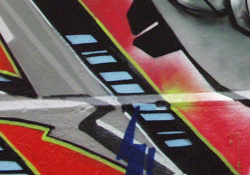}
  \includegraphics[width=1.1cm]{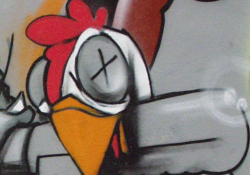}
  \includegraphics[width=1.1cm]{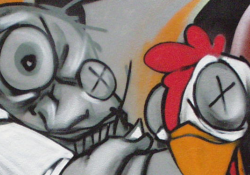}
  \includegraphics[width=1.1cm]{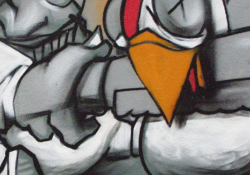}
  \includegraphics[width=1.1cm]{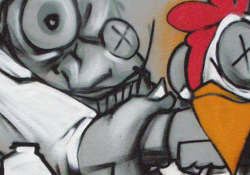}
  \includegraphics[width=1.1cm]{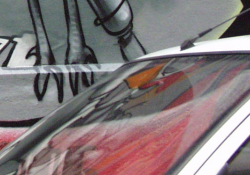}
  \includegraphics[width=1.1cm]{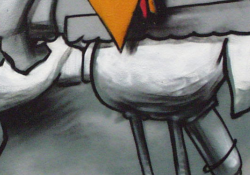}
  \includegraphics[width=1.1cm]{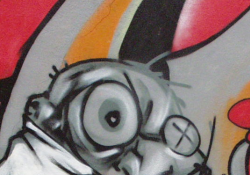}\\[0.1cm]
  \includegraphics[width=2.4cm]{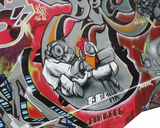}
  \includegraphics[width=2.4cm]{experimental-results-random-scene-2}
  \includegraphics[width=2.4cm]{experimental-results-random-scene-3}
  \includegraphics[width=2.4cm]{experimental-results-random-scene-4}
  \includegraphics[width=2.4cm]{experimental-results-random-scene-5}
  \caption{\label{fig:experimental-results-random-dataset}Models and
    scenes for the subset graf in the second experiment.}
\end{figure}

Because of its balance between accuracy and efficiency, circuits with
three vertices were chosen to represent keygraphs in the second
experiment. In this experiment, the objective was comparing
exhaustively the keygraph framework with the keypoint framework.  We
kept the previous dataset and the idea of cropping, but used all~$5$
degrees of all~$8$ distortions and instead of cropping manually we
randomly cropped~$10$ models for each
subset. Figure~\ref{fig:experimental-results-random-dataset} shows the
models and scenes for one of the subsets.

\begin{figure}[!ht]
  \centering
  \begin{tabular}{cc}
  \includegraphics[scale=0.3]{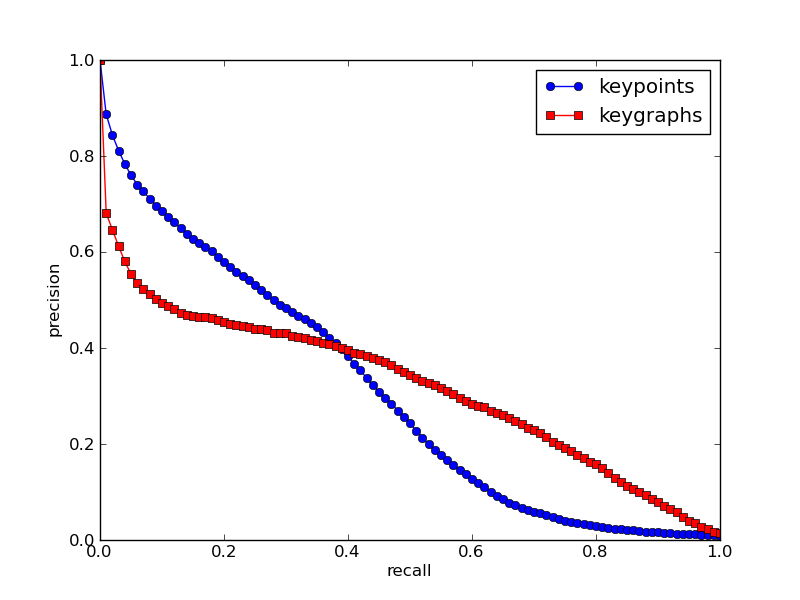}&
  \includegraphics[scale=0.3]{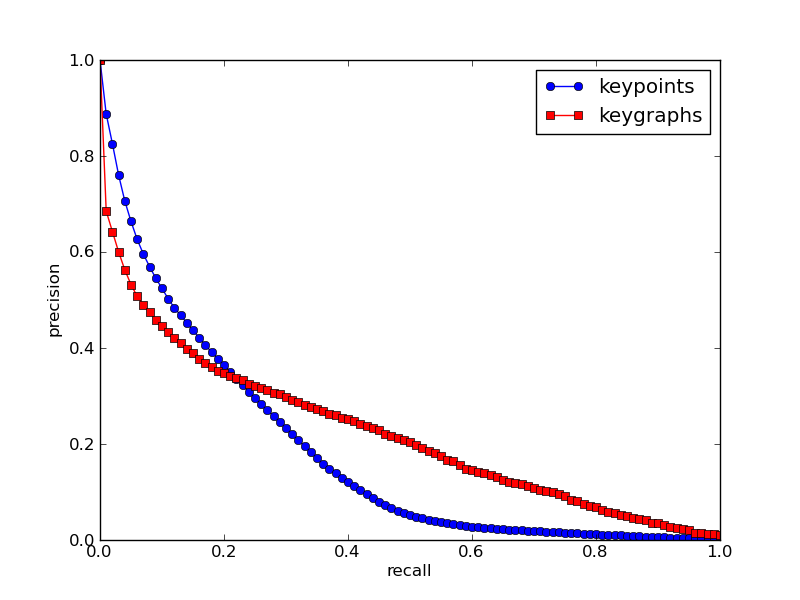}\\
  (a) Hessian-Affine & (b) MSER
  \end{tabular}
  \caption{\label{fig:experimental-results-random-all}Results for the
    two sequences of measurements in the second experiment: (a) for
    Hessian-Affine and (b) for MSER. Each curve is the mean of
    the~$200$ measurements taken for each keypoint detector.}
\end{figure}

\begin{figure}[!ht]
  \centering
  \begin{tabular}{cc}
  \includegraphics[scale=0.3]{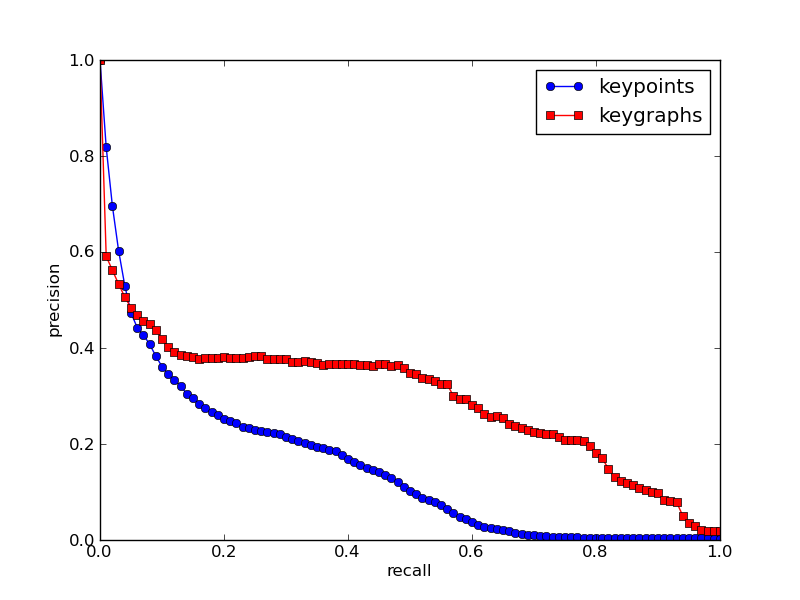}&
  \includegraphics[scale=0.3]{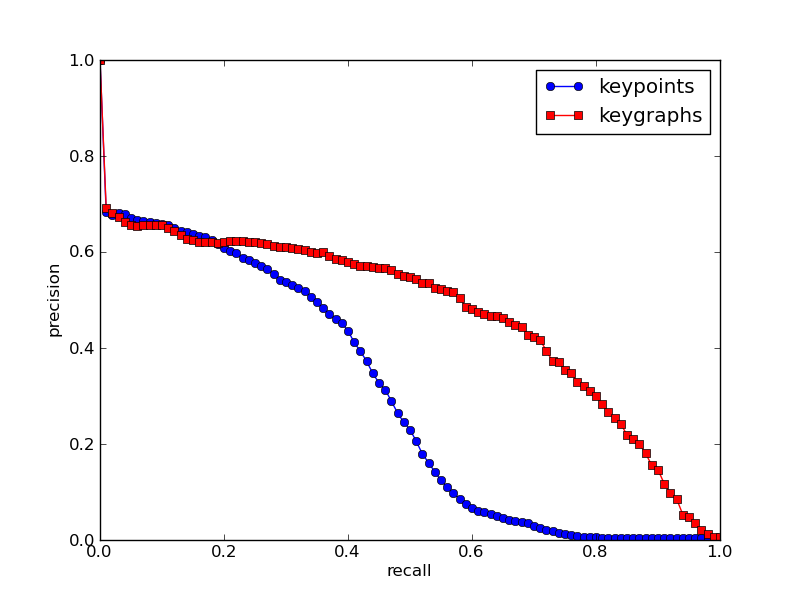}\\
  (a) graf & (b) ubc
  \end{tabular}
  \caption{\label{fig:experimental-results-random-precision-best}Results
    for the second experiment, specifically for Hessian-Affine and two
    subsets: (a) for graf and (b) for ubc. Each curve is the mean
    of~$50$ measurements.}
\end{figure}

\begin{figure}[!ht]
  \centering
  \begin{tabular}{cc}
  \includegraphics[scale=0.3]{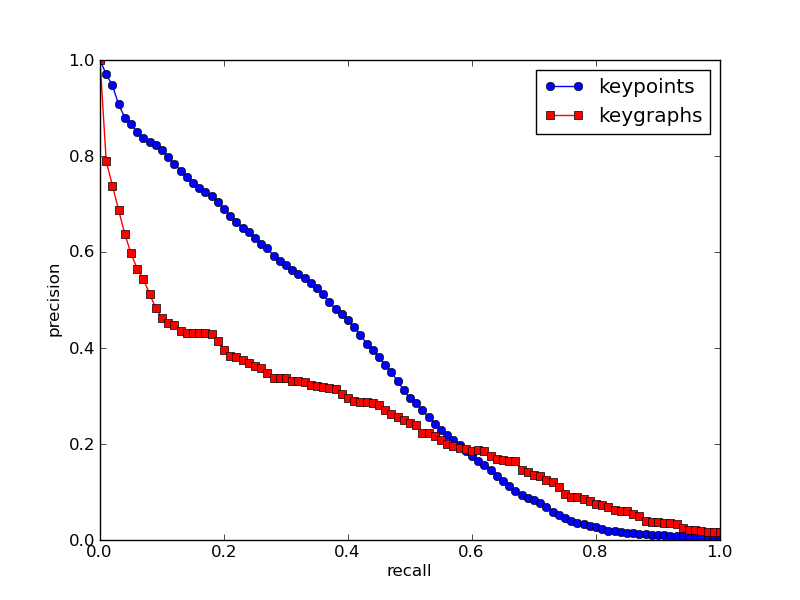}&
  \includegraphics[scale=0.3]{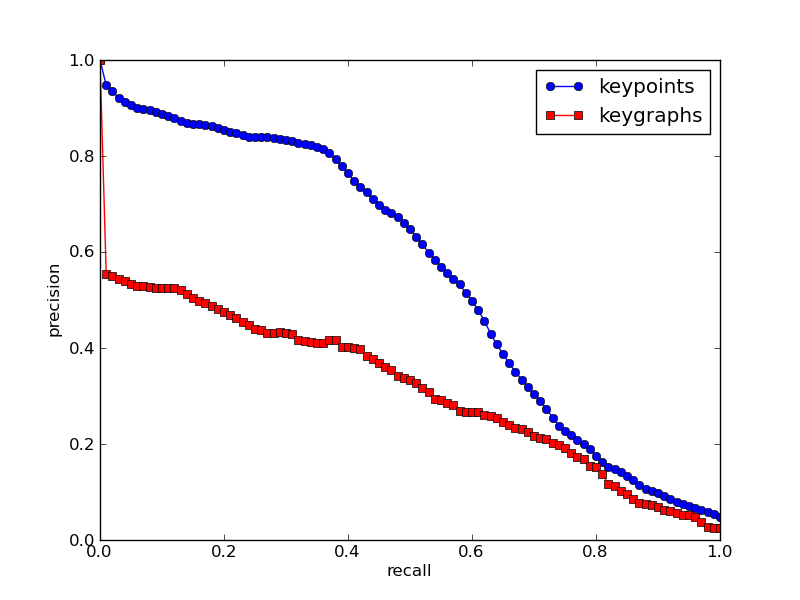}\\
  (a) bikes & (b) leuven
  \end{tabular}
  \caption{\label{fig:experimental-results-random-precision-worst-hessian}Results
    for the second experiment, specifically for Hessian-Affine and two
    subsets: (a) for bikes and (b) for leuven. Each curve is the mean
    of~$50$ measurements.}
\end{figure}

\begin{figure}[!ht]
  \centering
  \begin{tabular}{cc}
  \includegraphics[scale=0.3]{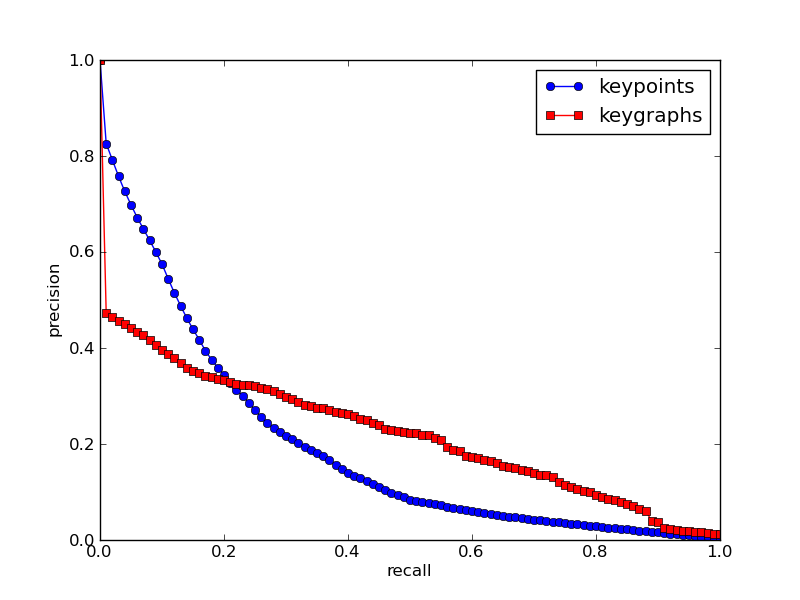}&
  \includegraphics[scale=0.3]{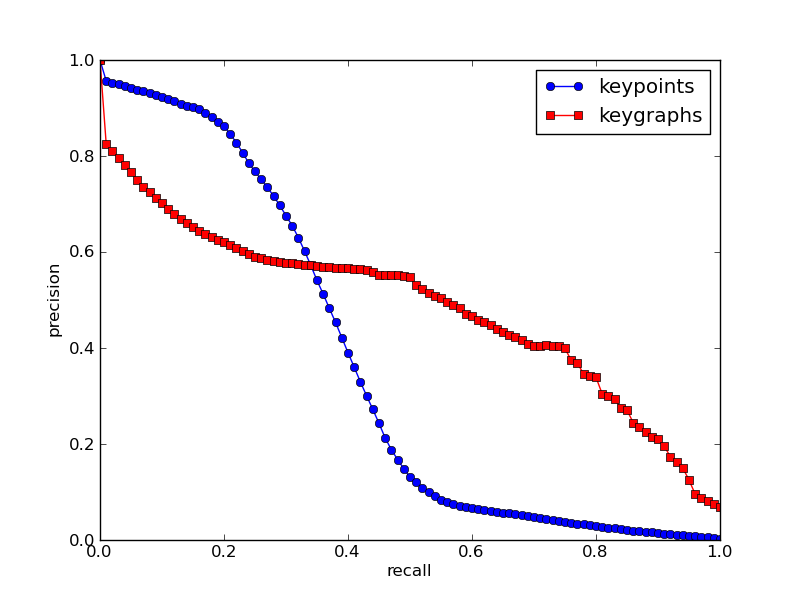}\\
  (a) bikes & (b) leuven
  \end{tabular}
  \caption{\label{fig:experimental-results-random-precision-worst-mser}Results
    for the second experiment, specifically for MSER and two subsets:
    (a) for bikes and (b) for leuven. Each curve is the mean of~$50$
    measurements.}
\end{figure}

Furthermore, we made a sequence of measurements with the
Hessian-Affine detector and a sequence of measurements with the MSER
detector. In total, $800$ measurements were made and their results are
shown in Figure~\ref{fig:experimental-results-random-all}. This second
experiment confirmed that keygraphs provide superior precision for
higher recalls. However, the specific recall where the curves
intersect each other was not stable across subsets and
detectors. Figure~\ref{fig:experimental-results-random-precision-best}
shows that for two subsets keygraphs performed consistently better
with Hessian-Affine. In contrast,
Figure~\ref{fig:experimental-results-random-precision-worst-hessian}
shows that for two subsets keygraphs performed consistently worse with
Hessian-Affine. With MSER the results for the same subsets were more
similar to the average results, as illustrated by
Figure~\ref{fig:experimental-results-random-precision-worst-mser}.

These differences are particularly interesting because they are direct
consequences of the advantages of each keypoint detector. In the
comparative study of
Mikolajczyk~et~al.~\cite{mtszmskv-2005:ijcv-65-1-43}, Hessian-Affine
performed worse precisely on the bikes and leuven subsets. This shows
that the keygraph framework is directly influenced by the
repeatability of the keypoint detector chosen, but when good
repeatability is ensured, then keygraphs present better precision at
higher recalls.

\begin{figure}[!ht]
  \centering
  \begin{tabular}{cc}
  \includegraphics[scale=0.3]{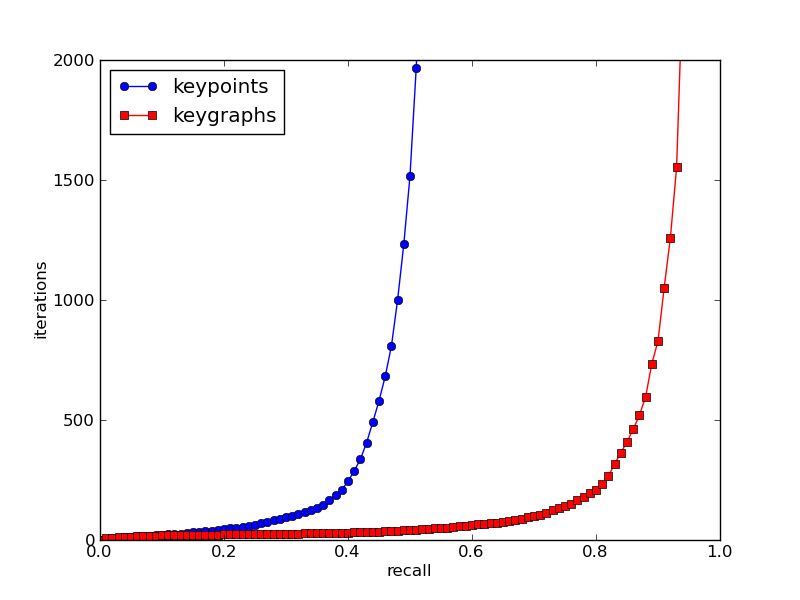}&
  \includegraphics[scale=0.3]{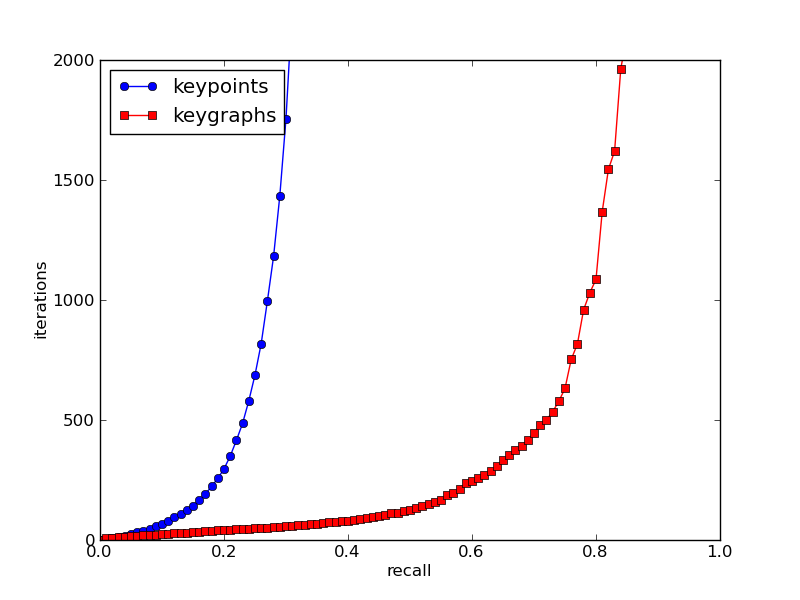}\\
  (a) Hessian-Affine & (b) MSER
  \end{tabular}
  \caption{\label{fig:experimental-results-random-iterations-all}Results
    for the two sequences of measurements in the second experiment:
    (a) for Hessian-Affine and (b) for MSER. Each curve is the mean of
    the~$200$ measurements taken for each keypoint detector.}
\end{figure}

\begin{figure}[!ht]
  \centering
  \begin{tabular}{cc}
  \includegraphics[scale=0.3]{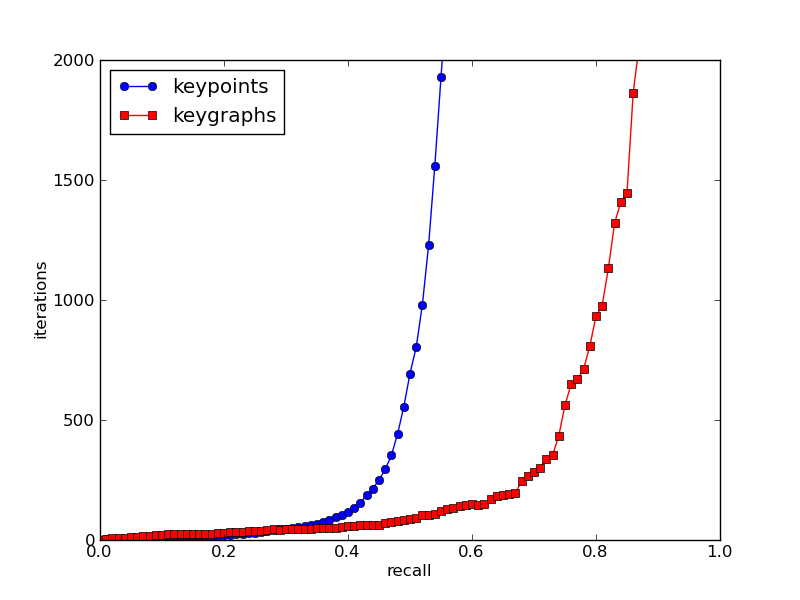}&
  \includegraphics[scale=0.3]{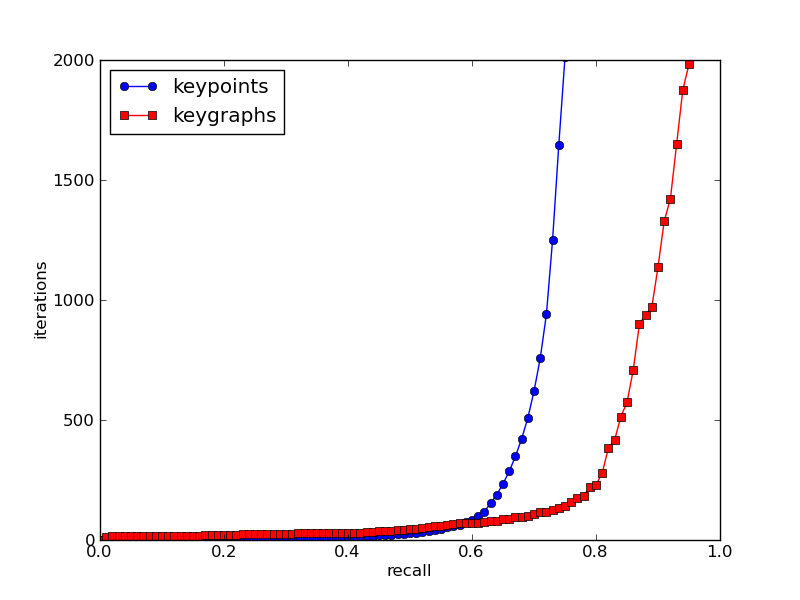}\\
  (a) bikes & (b) leuven
  \end{tabular}
  \caption{\label{fig:experimental-results-random-iterations-worst-hessian}Results
    for the second experiment, specifically for Hessian-Affine and two
    subsets: (a) for bikes and (b) for leuven. Each curve is the mean
    of~$50$ measurements.}
\end{figure}

But that does not mean that detectors with worse repeatability should
be necessarily dismissed. When using RANSAC, more important than the
precision~$p$ is the expected number of iterations, that can be
approximated by
\begin{displaymath}
  \frac{\log(1 - \prob)}{\log(1 - p^h)}
\end{displaymath}
where~$h$ is the size of the set of correspondences necessary for
estimating a pose candidate. The value of~$h$ is the fundamental
difference between keypoints and keygraphs: for estimating an
homography, four keypoint correspondences are needed. Therefore, $h =
4$ for keypoints and~$h = 2$ for circuits with three vertices. The
difference in number of iterations can be seen in
Figure~\ref{fig:experimental-results-random-iterations-all}. As can be
seen in
Figure~\ref{fig:experimental-results-random-iterations-worst-hessian},
even for the subsets where keygraphs had a consistently worse
precision, they still had a consistently better number of iterations.

\begin{figure}[!ht]
  \centering
  \includegraphics[width=2.40cm]{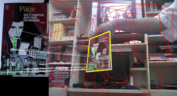}
  \includegraphics[width=2.40cm]{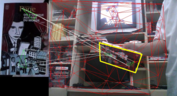}
  \includegraphics[width=2.40cm]{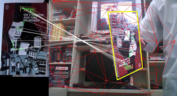}
  \includegraphics[width=2.40cm]{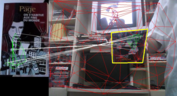}
  \includegraphics[width=2.40cm]{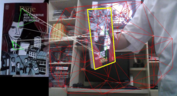}\\[0.1cm]
  \includegraphics[width=2.40cm]{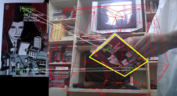}
  \includegraphics[width=2.40cm]{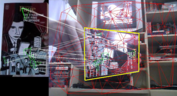}
  \includegraphics[width=2.40cm]{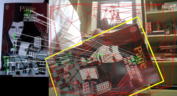}
  \includegraphics[width=2.40cm]{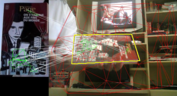}
  \includegraphics[width=2.40cm]{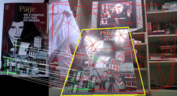}
  \caption{\label{fig:experimental-results-tracking-webcam}Screenshots
    of object detection with a fixed laptop webcam.}
\end{figure}

\begin{figure}[!ht]
  \centering
  \includegraphics[width=2.40cm]{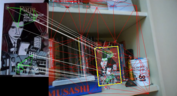}
  \includegraphics[width=2.40cm]{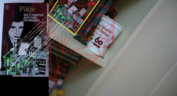}
  \includegraphics[width=2.40cm]{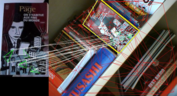}
  \includegraphics[width=2.40cm]{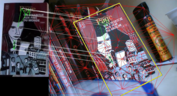}
  \includegraphics[width=2.40cm]{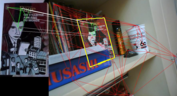}\\[0.1cm]
  \includegraphics[width=2.40cm]{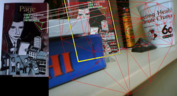}
  \includegraphics[width=2.40cm]{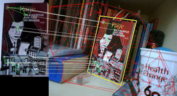}
  \includegraphics[width=2.40cm]{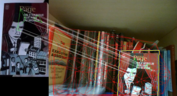}
  \includegraphics[width=2.40cm]{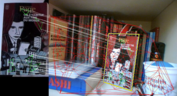}
  \includegraphics[width=2.40cm]{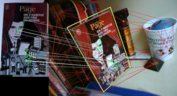} 
  \caption{\label{fig:experimental-results-tracking-eyetoy}Screenshots
    of object detection with a PlayStation Eye camera.}
\end{figure}

The objective of the third and final experiment was evaluating the
running time of the keygraph framework and its viability for real-time
applications. We wanted to verify if the implementation was capable of
detecting an object on each frame given by a camera in reasonable
time. Two sequences of tests were made: one with a fixed laptop
webcam, whose screenshots are shown in
Figure~\ref{fig:experimental-results-tracking-webcam}, and one with a
PlayStation Eye camera, whose screenshots are shown in
Figure~\ref{fig:experimental-results-tracking-eyetoy}.

Both cameras were configured to give~$640 \times 480$ frames and all
testes were made in a laptop with a Intel Core i5. Since the version
of OpenCV used was not compiled with support to parallelism, the
processing was made by a single 2.0 GHz core. For indexing the
keygraphs, we indexed the Fourier arc descriptors using the FLANN
library~\cite{ml-2009:visapp4-1-331} and for each scene
keytuple~$(a_s, b_s, c_S)$ we adopted the following procedure:

\small
\begin{center}
  \begin{tabular}{rl}
     1 & \hspace{0.0cm} let~$\Acal_m$ be the model arcs similar to~$(a_s, b_s)$\\
     2 & \hspace{0.0cm} let~$\Bcal_m$ be the model arcs similar to~$(b_s, c_s)$\\
     3 & \hspace{0.0cm} for each~$(a_m, b_m)$ in~$\Acal_m$\\
     4 & \hspace{0.5cm}   for each~$(b'_m, c_m)$ in~$\Bcal_m$ such that~$b'_m = b_m$\\
     5 & \hspace{1.0cm}     if~$(c_m, a_m)$ and~$(c_s, a_s)$ are similar\\
     6 & \hspace{1.5cm}       if~$(a_m, b_m, c_m)$ is the keytuple of a model keygraph\\
     7 & \hspace{2.0cm}         select the correspondence implied by~$(a_m, b_m, c_m)$ and~$(a_s, b_s, c_s)$\\
  \end{tabular}
\end{center}
\normalsize

The threshold applied to the Euclidean distance of Fourier descriptors
to decide whether two arcs were similar was~$0.5$. For both tests, the
average time for processing each frame was below~$30$ms, which ensures
at least~$30$ frames per second.  Using classic tracking techniques
such as statistical filtering would certainly improve even further.

\section{Conclusion}
\label{sec:conclusion}

We have proposed a framework for object detection based on a
generalization of keypoint correspondences into keygraph
correspondences. The main difference of our approach to similar ones
in the literature is that each step deals directly with graphs: the
description does not need point region descriptors, the comparison
takes the entire structure into consideration, and the pose estimation
uses graph correspondences directly instead of converting them with a
voting table. In order to show the viability of the framework, we also
proposed an implementation with low theoretical complexity and
performed experiments to prove both its accuracy and speed.

Besides testing different methods for keygraph detection and
description, an interesting possibility for future research is
developing keypoint detectors specifically tailored for building good
keygraphs. A good step in this direction are the critical nets
recently proposed by
Gu~et~al.~\cite{gzt-2010:eccv11iii-lncs-6313-663}. Another important
step is showing that relatively powerful computers are not needed for
applications based on the framework. Our implementation has been
recently ported to a mobile phone, much more limited than the laptop
used in the experiments of this paper, and had success in indoor sign
detection~\cite{mhpch-2011:gbr8-lncs-6658-315}.

\section*{Acknowledgments}

This research is supported by FAPESP (processes 2007/56288-1 and
2011/50761-2), CAPES, CNPq and FINEP. The authors would like to thank
Roberto Hirata Jr., Carlos Hitoshi Morimoto, Ana Beatriz Vicentim
Graciano, Alexandre Noma, Jesus Mena-Chalco, Henrique Morimitsu,
Ricardo da Silva Torres, Eduardo Valle, Isabelle Bloch, Geoffroy
Fouquier and Valerie Gouet-Brunet for valuable help and
suggestions. 
We are grateful to the reviewers who helped to greatly improve the
paper.

\bibliographystyle{amsplain}

\bibliography{arxiv}

\end{document}